\pdfoutput=1
\documentclass[letterpaper,10pt]{article}
\usepackage[table]{xcolor}
\usepackage{times}
\usepackage{tabularx} 
\usepackage{amsmath}  
\usepackage{graphicx} 
\usepackage[margin=1in]{geometry} 
\usepackage{cite} 
\usepackage[final]{hyperref} 
\hypersetup{
	colorlinks=true,       
	linkcolor=blue,        
	citecolor=blue,        
	filecolor=magenta,     
	urlcolor=blue         
}
\usepackage{natbib}
\usepackage[utf8]{inputenc} 
\usepackage[T1]{fontenc}    
\usepackage{hyperref}
\usepackage{url}            
\usepackage{booktabs}       
\usepackage{amsfonts}       
\usepackage{nicefrac}       
\usepackage{microtype}      

\usepackage{xspace}  
\usepackage{cleveref}
\usepackage[ruled,vlined]{algorithm2e}
\usepackage{worldflags}
\usepackage{tikz}
\usepackage{enumitem}
\usepackage{wrapfig}
\usepackage{multicol}
\usepackage{multirow}
\newcommand{\llms}{LLMs\xspace}
\usepackage{todonotes}
\usepackage{makecell}
\usepackage{bbding}
\usepackage{adjustbox}
\usepackage{subfigure}
\usepackage{caption}
\usepackage{etoc}
\usepackage{blindtext}

\usepackage{subfigure}
\usepackage{dsfont}
\usepackage{amssymb}  
\usepackage{makecell}
\usepackage{amsmath}
\usepackage{multirow}
\usepackage{booktabs}

\usepackage{algpseudocode}
\usepackage{amsthm}
\usepackage{xspace}
\usepackage{longtable}
\usepackage{wrapfig}
\usepackage{enumitem}
\usepackage{etoc}
\usepackage{worldflags}
\usepackage{tikz}

\usepackage{xspace}

\newcommand{\prompt}[1]{{\small \ttfamily #1}\xspace}
\newcommand{\method}{MentalArena\xspace}
\newcommand{\encode}{\textit{Symptom Encoder}\xspace}
\newcommand{\decode}{\textit{Symptom Decoder}\xspace}
\newcommand{\optimize}{\textit{Model Optimizer}\xspace}

\usepackage{todonotes}

\begin{document}
\title{\method: Self-play Training of Language Models for Diagnosis and Treatment of Mental Health Disorders}

\author{Cheng Li$^{1,2}$\footnote{Work done during Cheng's internship at University of Illinois Urbana-Champaign. Contact: chenglicat0228@gmail.com.}, May Fung$^{1}$, Qingyun Wang$^1$, Chi Han$^1$, Manling Li$^3$, Jindong Wang$^{2}$$^\dagger$, Heng Ji$^1$\footnote{Corresponding authors.}\\
$^{1}$University of Illinois Urbana-Champaign \\ $^{2}$Microsoft Research Asia\\ $^3$Stanford University
}


\date{}
\etocdepthtag.toc{mtchapter}
\etocsettagdepth{mtchapter}{none}
\etocsettagdepth{mtappendix}{none}
\maketitle

\begin{abstract}
Mental health disorders are one of the most serious diseases in the world. 
Most people with such a disease lack access to adequate care, which highlights the importance of training models for the diagnosis and treatment of mental health disorders. However, in the mental health domain, privacy concerns limit the accessibility of personalized treatment data, making it challenging to build powerful models.
In this paper, we introduce \emph{\method}, a self-play framework to train language models by generating domain-specific personalized data, where we obtain a better model capable of making a personalized diagnosis and treatment (as a therapist) and providing health information (as a patient). 
To accurately model human-like mental health patients, we devise \encode, which simulates a real patient from both cognition and behavior perspectives. \decode simulates the interactions between a mental health patient and a therapist for diagnosis and treatment, generating more personalized dialogues while mitigating cognitive bias.
We evaluated \method against $8$ benchmarks, including biomedicalQA and mental health tasks, compared to $6$ advanced models. Our models, fine-tuned on both GPT-3.5 and Llama-3-8b, significantly outperform their counterparts, including GPT-4o. We hope that our work can inspire future research on personalized care. 
\end{abstract}

\section{Introduction}

Mental health disorders include a variety of conditions such as anxiety, depression, and schizophrenia, which affect people's thinking, emotions, behavior, or mood~\citep{prince2007no}.
In 2019, approximately $970$ million people worldwide lived with a mental health disorder, with anxiety and depression being most prevalent~\citep{who}.
The number increased by $28\%$ in 2020 and continues to increase.
Despite the availability of effective treatments, many individuals lack access to adequate care due to under-resourced health systems.
For example, only 29\% of people with psychosis and one third of people with depression receive formal mental healthcare~\citep{who}.
It is indispensable to develop machine learning models for the assistance of diagnosis and treatment for such diseases.
However, existing AI therapist systems use templates and decision trees, which are not flexible to meet the large demands on personalized care~\citep{fiske2019your,d2020ai,grodniewicz2023waiting,devarakonda2019flare}.

\begin{figure*}[t!]
    \centering
    \includegraphics[width=\textwidth]{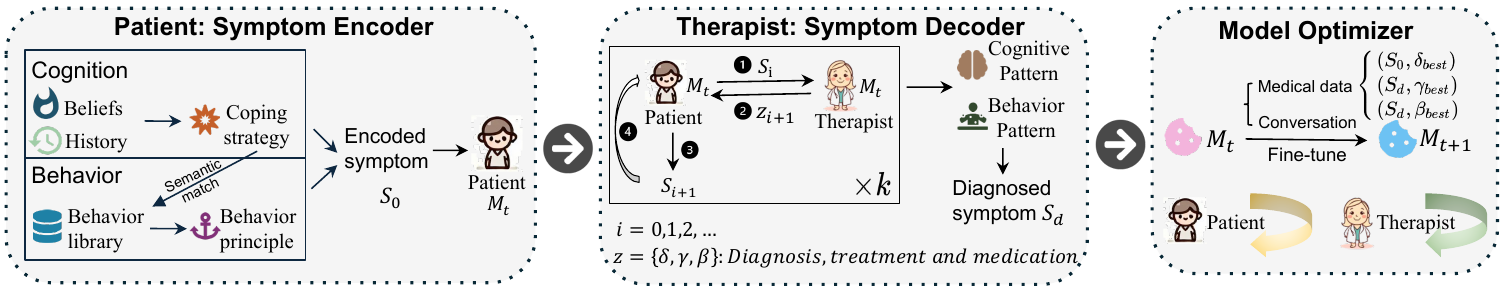}
    \caption{\method is a self-play framework for assistance in the diagnosis and treatment of mental health disorder consisting of three modules: \encode, \decode, and \optimize.
    }
    \label{fig-overview}
\end{figure*}

One of the keys to training powerful models lies in collecting high-quality training data from the real world. 
However, due to privacy concerns in the medical domain, data collection, especially personalized data for mental health disorders, is inherently challenging. 
A growing body of work has focused on enhancing mental health language models by sourcing additional domain-specific data from social media~\citep{xu2024mental, yang2024mentallama, hu2024psycollm}. 
However, social media data is inherently biased and under-representative, failing to capture the full spectrum of individuals' mental health needs. 
Furthermore, as \llms continue to scale, the availability of real-world training data becomes increasingly limited, exacerbating this challenge.


Recently, several efforts have focused on self-play~\citep{hu2024agentgen, yang2024collaborative, liang2024sheep, wu2024meta, unleashing2024}, where models play different roles and self-evolve or co-evolve during interactions with other models. 
A model generates training data independently and then uses this synthetic data to train itself. However, there are two key challenges that prevent us from adopting self-play training for mental health disorders: 1) Scarcity of high-quality data: Mental health disorders are complex, involving cognitive and behavioral symptoms. Current \llms lack the personalized experience needed to accurately simulate patients with diverse conditions~\citep{schmidgall2024agentclinic, wang2024patient}. 
2) Cognitive bias: Cognitive bias often arises, where the therapist misinterprets the patient's intent due to knowledge gaps and insufficient communication skills, and the patients are hesitant to express their symptoms and feelings. This mirrors real-world therapist-patient misunderstandings~\citep{britten2000misunderstandings, west1984medical, shreevastava2021detecting}.

\begin{figure*}[t!]
    \centering
    \includegraphics[width=\textwidth]{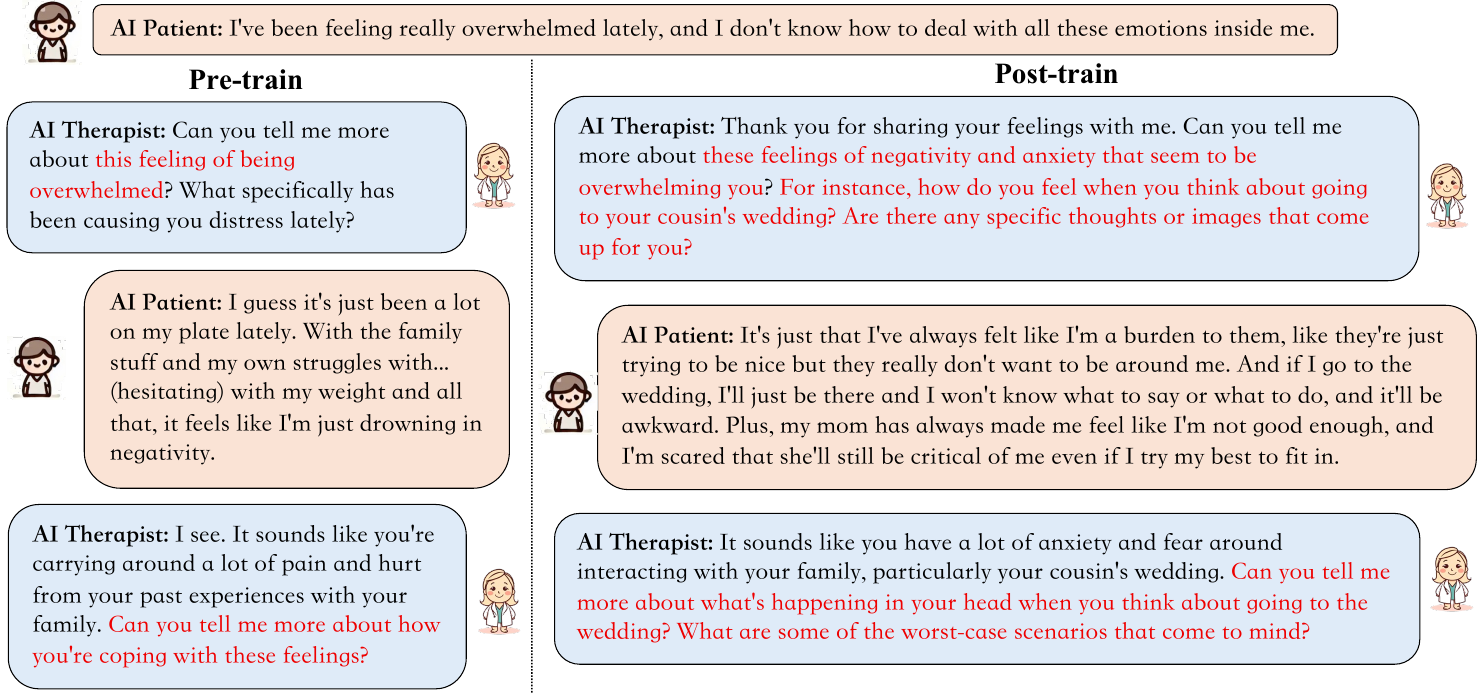}
    \caption{Case study on pre-training and post-training.
    }
    \label{fig-example}
\end{figure*}

\method is a framework specifically designed for self-play training of language models to assist in the diagnosis, treatment, and medication of mental health disorders.\footnote{\method is intended to assist in the diagnosis, treatment, and medication of mental health disorders, rather than train an independent AI-therapist. Training an independent AI-therapist is still a long-term goal.} 
The model $M$ plays dual roles: both the mental health patient and the therapist. As the therapist, it provides diagnoses, treatment plans, and medication regimens based on the symptoms presented by the patient. As the patient, it provides health information and simulates its updated health status after each treatment and medication plan is implemented.
As illustrated in \Cref{fig-overview}, \method comprises three key modules: \encode, \decode, and \optimize. 
\encode models mental health patients based on cognitive models\footnote{The cognitive model is designed based on cognitive behavioral therapy (CBT) principles~\citep{beck2020cognitive}, a well-established paradigm in psychotherapy.} 
and behavioral patterns, offering rich insights into coping strategies and behavioral principles. \decode simulates the interactions between a mental health patient and a therapist for diagnosis and treatment, generating more personalized dialogues while mitigating cognitive bias~\citep{britten2000misunderstandings, west1984medical}.
During each iteration, we collect data from these interactions, including diagnostic, treatment, and medication datasets, along with calibrated conversations via the \textit{cognitive bias check}. This data is used to evolve the models through training. \Cref{fig-example} presents examples of the pre-training and post-training results. We observe that the AI therapist demonstrates improved communication skills after training and is better able to guide the AI patient to provide more detailed descriptions of the symptoms.

We conduct experiments on eight benchmarks, including datasets for biomedical QA and mental health detection. We compare our fine-tuned models with other state-of-the-art models and specialized mental health models. 
Additionally, we benchmark against two advanced prompt engineering approaches. Our models outperform all counterparts, including GPT-4o~\citep{gpt4o}. Specifically, \method achieves significant improvements over the base models, with a 17.71\% increase in performance over GPT-3.5-turbo and a 6.86\% improvement over Llama-3-8b.

We further conduct a thorough analysis of the dynamics of self-play training. We find that the perplexity score~\citep{marion2023less, wang2023making} is highly correlated with model performance. Regarding diversity gain~\citep{bilmes2022submodularity}, the model performance improves when the diversity gain exceeds certain thresholds.
Our user study indicates that our models significantly improve the ability to role-play both as patients and therapists. The experimental results also demonstrate that most of our synthetic datasets are valid. 
We also investigate whether \method can generalize to other medical domains. The results on MedMCQA~\citep{pal2022medmcqa} and MMLU~\citep{hendrycks2020measuring} demonstrate \method’s ability to generalize across various medical fields.
Additionally, we examine the issue of catastrophic forgetting in our fine-tuned models. Results from $21$ tasks in BIG-Bench-Hard (BBH)~\citep{suzgun2022challenging} show that our models maintain or even improve performance on general benchmarks.

In summary, the contributions of this paper are as follows: 
\begin{enumerate}[leftmargin=1em]
\setlength\itemsep{0em}
    \item We propose \method, a novel and cost-effective self-play framework for training language models to assist in diagnosing and treating mental health disorders.\footnote{\method trains models using self-synthesized data rather than real-world data, making it more cost-effective and addressing data privacy concerns.}\method introduces \encode and \decode, designed to simulate real patient-therapist interactions by modeling cognitive and behavioral processes. 
    \item Using \method, we generate high-quality data containing diagnosis, treatment, and medication data. There are $18k$ samples in total that can be used for further training and research.
    \item We evaluate \method on $8$ benchmarks comparing with $6$ \llms. Our models based on GPT-3.5-turbo and Llama-3-8b that are trained through the \method framework outperform all off-the-shelf counterparts, including GPT-4o.
\end{enumerate}

\section{Related Work}

\subsection{Large Language Models for healthcare}

Researchers have explored the potential of large language models (LLMs) in healthcare~\citep{jiang2023health, li2023chatdoctor,liu2023large,lupetti2023trustworthy,nori2023capabilities,singhal2023towards,wu2023pmc,wang2024rulealign,li2023good}. 
In the mental health domain, research has taken two main approaches. The first involves fine-tuning domain-specific \llms on existing datasets or social media data~\citep{yang2024mentallama,xu2024mental}. The second approach enhances mental health performance through prompt engineering~\citep{yang2023towards}.

Unlike previous methods, \method fine-tunes mental health models through self-play training, in which the base model assumes both patient and therapist. 

\subsection{Self-play frameworks in Large Language Models}

Self-play involves a model evolving through interactions with copies of itself, creating a feedback loop that refines performance without external input. It is particularly effective in environments where the model simulates multiple roles, such as multiplayer games~\citep{silver2016mastering,silver2017mastering}. Compared to interactive methods, self-play provides a more efficient strategy for obtaining feedback without relying on an external environment~\citep{taubenfeld2024systematic,ulmer2024bootstrapping,lu2024large,askari2024self}.


To overcome the lack of patient data in training corpus, \method introduces \encode, a component designed to effectively model real mental health patients.

\section{\method}
\label{sec-method}

\begin{figure*}[t!]
    \centering
    \includegraphics[width=.95\textwidth]{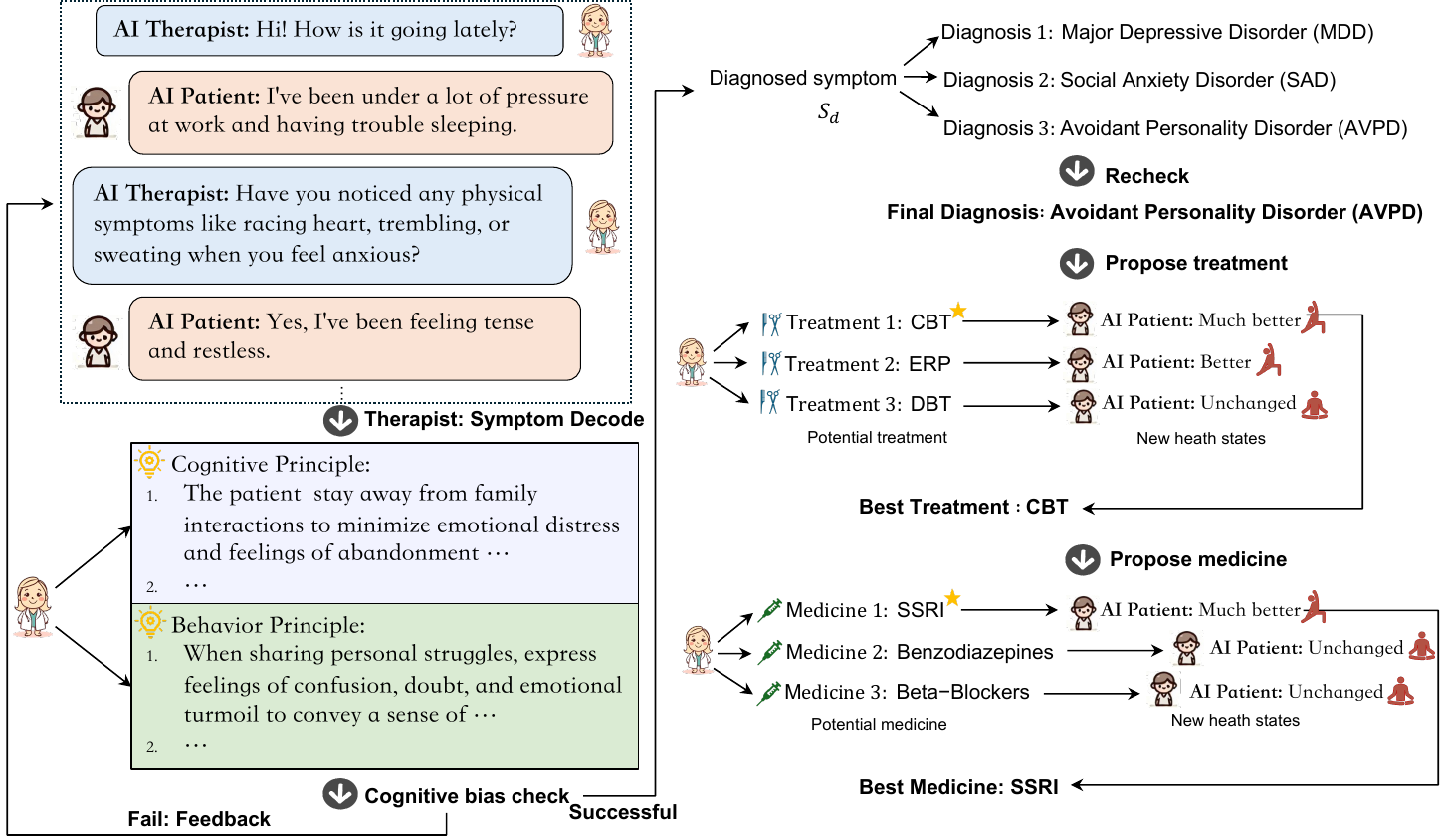}
    \vspace{-.1in}
    \caption{
    Symptom Decoder simulates the diagnostic and therapeutic interactions between patients and therapists, enabling the generation of personalized dialogues while addressing cognitive bias.
    }
    \label{fig-method}
    \vspace{-.1in}
\end{figure*}

\subsection{Preliminaries}

We first review the process of diagnosing and treating mental health disorders, highlighting key concepts.
The diagnosis of mental health conditions begins with an assessment of an individual's \textit{health state}, which encompasses their mental and emotional well-being. \textit{Symptoms} serve as critical indicators of potential issues and may include emotional changes (e.g., anxiety, depression), cognitive impairments (e.g., memory problems), and behavioral changes (e.g., social withdrawal).
These symptoms inform a formal \textit{diagnosis}, typically determined through clinical interviews to identify specific disorders such as depression, anxiety, or schizophrenia.\footnote{In our framework, diagnosis is based solely on clinical interviews due to limited data availability for laboratory tests.}
Following a diagnosis, the \textit{treatment} process begins, often involving a combination of psychotherapy (e.g., cognitive-behavioral therapy), lifestyle modifications, and, in some cases, medication. \textit{Medications}, such as antidepressants and mood stabilizers, are prescribed to regulate brain chemistry and alleviate symptoms.~\citep{prince2007no}


\subsection{Overview of the Framework}

Several studies have explored self-play training across various scenarios and tasks~\citep{taubenfeld2024systematic,ulmer2024bootstrapping,lu2024large,askari2024self,wang2024sotopia,unleashing2024,chen2024self}. However, its application in the medical domain remains both underexplored and challenging due to two primary issues: data deficiency in medical contexts and the problem of cognitive bias between patients and therapists.
The first issue, data deficiency, complicates the simulation of a patient role~\citep{schmidgall2024agentclinic,wang2024patient,yu2024aipatient}, as training corpora often lack sufficient patient-specific data. The second issue, cognitive bias, hinders effective diagnosis and treatment by introducing ambiguity into symptom interpretation and communication.
To address these challenges, we propose generating medical data to mitigate knowledge gaps, enabling more accurate role-playing as both therapists and mental health patients. Additionally, we produce high-quality conversations calibrated via an \textit{cognitive bias check}, which improves therapists' communication skills and mitigating misdiagnoses and inappropriate treatment suggestions.

\method is a framework specifically designed for the self-play training of language models to support the diagnosis, treatment, and medication management of mental health disorders.
As illustrated in \Cref{fig-overview}, \method comprises three key modules: \encode, \decode, and \optimize.
The \encode module models mental health patients by leveraging cognitive models and behavioral patterns, providing rich insights into coping strategies and behavioral principles.
The \decode module simulates the diagnostic and therapeutic interactions between patients and therapists, enabling the generation of personalized dialogues while addressing cognitive bias~\citep{britten2000misunderstandings,west1984medical}.
During each iteration, we collect data generated from interactions, including medical data and calibrated conversations between patients and therapists. These datasets are then used to iteratively train the models for performance enhancement.

Formally, let $\mathbf{S}$ represent the health state of a patient, which can also be interpreted as the encoded symptoms.
We denote the base model (e.g., GPT-3.5) as $M$, which is used to simulate both the therapist and patient via a role-play strategy.
In iteration $t$, the model $M_t$ plays the therapist $D_t=M_t(\cdot \mid \text{Prompt}_{doc})$ and the patient $P_t=M_t(\cdot \mid \text{Prompt}_{pat})$.
Our objective is to derive an optimized model $M^\ast$ by self-play training. This model should achieve superior performance in both personalized diagnosis and treatment (as the therapist) and effective information disclosure (as the patient).

\subsection{Patient: Symptom Encoder}
\label{sec-encode}

The \encode module mimics mental health patients from both cognitive and behavioral perspectives, generating meaningful symptoms $S_0$ by leveraging cognitive models and behavioral principles.
The cognitive model is grounded in the principles of cognitive behavioral therapy (CBT)~\citep{beck2020cognitive}, a widely used paradigm in psychotherapy.
A detailed explanation of cognitive model can be found in \Cref{sec-append-cog-intro}.
We utilize $106$ patient cognitive models derived from prior work~\citep{wang2024patient}, which were developed by clinical psychologists. To simulate the cognitive activity of mental health patients, we encode these cognitive models into patient representations using customized prompts. Examples of these prompts are provided in \Cref{sec-append-prompt}.

For patient behavior modeling, we use behavior principles compiled by \cite{louie2024roleplay} as a behavior library, developed by $25$ mental health experts. Examples of behavior patterns can be found in \Cref{sec-append-seed-example}.
To identify the most suitable behavior pattern for each cognitive model, we begin by semantically matching the coping strategies of the cognitive model with each behavior pattern in the library. We use Bert-base~\citep{devlin2018bert} to generate embeddings for both coping strategies and behavior principles, balancing effectiveness with computational cost.
Semantic similarity scores are calculated between the coping strategies and each behavior pattern. The maximum similarity score across all behavior principles within a given behavior pattern is used to represent the pattern's overall score. The five behavior patterns with the highest scores are selected for further evaluation.
To ensure the selection of the most appropriate pattern, we prompt GPT-4-turbo~\citep{openai2023gpt4} to choose one from the top five patterns. The finalized behavior pattern is then integrated into the patient representation via prompt, as detailed in \Cref{sec-append-prompt}.

\subsection{Therapist: Symptom Decoder}
\label{sec-decode}

Mental health patients are often reluctant to discuss their symptoms openly with therapists. During interactions between a real therapist and patient, misunderstandings can arise when the patient attempts to express an opinion, but the therapist misinterprets the intent due to knowledge gaps or insufficient communication skills~\citep{britten2000misunderstandings,west1984medical}.
A similar issue, termed cognitive bias, can occur in conversations between AI-simulated therapists and patients, leading to inaccurate diagnoses and treatments. To address this, the \decode module is designed to calibrate therapist-patient conversations and mitigate cognitive bias effectively.
Following several interactions, the therapist reviews the patient’s health information from previous conversations and performs an in-depth analysis of the patient’s cognitive and behavioral patterns, resulting in a diagnosed symptom state $S_d$. Then a \textit{ cognitive bias check} is performed by semantically matching the initial encoded symptom, $S_0$, with the diagnosed symptom, $S_d$. Discrepancies between $S_0$ and $S_d$ are used to provide suggestions for therapists and guide subsequent conversations, ensuring that the communication becomes progressively more accurate.
This iterative process continues until cognitive bias is eliminated, allowing the therapist to provide a precise diagnosis and appropriate treatment for the patient. The conversation will end if it cannot pass the \textit{cognitive bias check} five times.

As illustrated in \Cref{fig-method} (left), after multiple conversations, the therapist decodes the patient’s cognitive and behavioral principles. For instance, the decoded cognitive principle might be: ``The patient stay away from family interactions to minimize emotional distress and feelings of abandonment...'' Similarly, the decoded behavioral principle might state: ``When sharing personal struggles, express feelings of confusion, doubt, and emotional turmoil to convey a sense of vulnerability and authenticity.''
The semantic similarity score between the decoded symptom, $S_d$, and the encoded symptom, $S_0$, is then calculated. If the score exceeds 0.9, the conversation concludes, indicating that the therapist has fully understood the patient’s health state. Otherwise, a lower score suggests the presence of cognitive bias.
To improve therapist communication skills, the differences between $S_d$ and $S_0$ are summarized, and feedback is generated using GPT-4-turbo~\citep{openai2023gpt4}. This feedback highlights missing or unclear information about the patient, guiding the therapist in formulating further inquiries. For example, the feedback might suggest: ``The therapist could focus on exploring what has been causing the patient to feel tense.'' The conversation continues iteratively until the similarity score between $S_d$ and $S_0$ exceeds 0.9, ensuring the therapist gains an accurate understanding of the patient’s health state.

After the conversation concludes, the therapist analyzes the patient’s final diagnosed symptom, $S_d$, and formulates several diagnostic plans ($\delta_1$, $\delta_2$, ..., $\delta_n$). To ensure diagnostic accuracy, the patient reviews each plan and selects the most appropriate one based on their health condition. Subsequently, the therapist develops a series of treatment and medication plans ({$\gamma_1$, $\beta_1$}, ..., {$\gamma_k$, $\beta_k$}) aligned with the selected diagnosis ($\delta_{best}$).
To determine the optimal plan for treatment and medication, each plan taken by the patient denoted as the encoded symptom $S_0$, and the progression of the patient’s encoded symptoms is monitored. As treatment and medication plans ({$\gamma_1$, $\beta_1$}, ..., {$\gamma_k$, $\beta_k$}) are administered, the patient’s encoded symptom evolves, updating sequentially to $S_1$, $S_2$, ..., $S_k$, reflecting their changing health state.
As shown in the center of \Cref{fig-overview}, the patient first transmits $S_0$ to the therapist. Following the administration of the initial treatment or medication, $z_1$, the encoded symptom is updated to $S_1$. Similarly, after applying the subsequent intervention, $z_2$, the encoded symptom progresses to $S_2$.
Encoded symptoms act as indicators of the effectiveness of the treatment and medication plans, providing feedback on their impact as the interventions proceed. Ultimately, the therapist identifies and provides the optimal diagnostic, treatment, and medication information $\mathbf{z}_{best}={\delta_{best}, \beta_{best}, \gamma_{best}}$, which is essential for model optimization.

\subsection{Model Optimizer}
\label{sec-train}

After obtaining the treatment, diagnosis, and medication datasets through \decode, we train the model, $M$, in a self-play manner to enhance its ability to deliver personalized diagnoses and treatments (as a therapist) and correctly present information (as a patient).
The model is fine-tuned using paired data, such as $(S_0, \delta_{best})$, $(S_d, \gamma_{best})$, and $(S_d, \beta_{best})$, to address knowledge deficiencies. Examples are shown in \Cref{fig-train-data}. Additionally, calibrated conversations are incorporated into the training process to improve the therapist's communication skills, ensuring more accurate and empathetic interactions.
Detailed training settings are provided in \Cref{sec-append-training-details}.

\section{Experiment}

\begin{table*}[t!]
\caption{Main results on Accuracy (\%) for \method with different base models. The first five rows are either strong methods (i.e., GPT-4o) or those designed specifically for mental health.}
\label{tb-results-main}
\centering
\resizebox{.9\textwidth}{!}{
\begin{tabular}{cccccccccc}
\toprule
Model                & MedQA                & MedMCQA              & PubMedQA             & CAMS                 & dreaddit             & Irf                  & DR                   & T-SID                & AVG                  \\ \midrule
MentaLLaMa-13b       & 28.32                & 12.42                & 28.96                & \textbf{37.28}                & 62.08                & 46.81                & 75.56                & 61.99                & 44.18                \\
Mental-LLM-alpaca    & 28.32                & 12.42                & 0.00                 & 29.76                & \textbf{64.98}                & 51.96                & 75.80                & 53.49                & 39.59                \\
Mental-LLM-t5        & 0.00                 & 0.32                 & 49.09                & 27.04                & 63.29                & 47.70                & \textbf{76.05}                & 54.22                & 39.71                \\
GPT-4o               & 87.86                & 74.20                & 60.06                & 27.68                & 49.03                & \textbf{64.65}                & 25.52                & 61.00                & 56.25                \\
GPT-4o+MedPrompt     & \textbf{90.17}                & \textbf{78.34}                & \textbf{67.38}               & 31.52                & 53.27                & \textbf{64.65}                & 26.01                & \textbf{62.13}                & \textbf{59.18}                \\ \midrule
Base: GPT-3.5-turbo  & 64.16                & 33.76                & 44.68                & 28.96                & 46.03                & 64.65                & 23.15                & 59.96                & 45.67                \\
+Chain-of-thought     & 65.90                & 37.97                & 45.73                & 29.92                & 48.05                & 64.65                & 24.20                & 58.92                & 46.92                \\
+MedPrompt            & 69.94                & 43.89                & 47.26                & 30.2                 & 49.03                & 64.65                & 24.20                & 61.03                & 48.77                \\
\rowcolor[gray]{0.9} +Ours                 & \textbf{74.57}                & \textbf{91.08}                & \textbf{97.56}                & \textbf{97.56}                & \textbf{55.20}                & \textbf{64.65}                & \textbf{28.04}                & \textbf{63.10}               & \textbf{63.38}       \\ \midrule
Base: Llama-3-8b     & 70.52                & 42.04                & 86.59                & 25.12                & 58.45                & 45.76                & 74.81                & 48.80                & 56.51                \\
+Chain-of-thought     & 75.14                & 47.77                & 88.21                & 33.6                 & 62.22                & 45.91                & 75.80                & 53.03                & 60.21                \\
+MedPrompt            & 76.88                & 49.41                & 89.99                & \textbf{35.08}                & 61.59                & 48.05                & 76.03                & 57.65                & 61.83                \\
\rowcolor[gray]{0.9} +Ours                 & \textbf{78.03}                & \textbf{50.32}                & \textbf{92.68}                & 29.60                & \textbf{65.46}                & \textbf{52.25}                & \textbf{76.54}                & \textbf{76.54}                & \textbf{63.37}       \\ \bottomrule
\end{tabular}
}
\end{table*}

\subsection{Setup}


\textbf{Datasets:} As summarized in \Cref{tb-testset}, we adopt $8$ datasets: MedQA~\citep{jin2021disease}, MedMCQA~\citep{pal2022medmcqa}, PubMedQA~\citep{jin2019pubmedqa}, CASM~\citep{garg2022cams}, Dreaddit~\citep{turcan2019dreaddit}, Irf~\citep{garg2023annotated}, DR~\citep{pirina2018identifying} and T-SID~\citep{ji2022suicidal}.
Our evaluation spans biomedical QA and mental health detection, covering knowledge on diagnosis, treatment, and medication.
These datasets include general mental health tasks, such as depression/suicide, stress, and interpersonal risk factors detection, as well as real-world mental health cases. Details are provided in \Cref{sec-append-benchmark}

\textbf{Baselines:} We compare our models to other mental health models by different prompt engineering methods.
For baseline models, we compare with the state-of-the-art \llms: GPT-3.5-turbo~\citep{chatgpt}, GPT-4o~\citep{gpt4o} and Llama-3-8b~\citep{dubey2024llama}. We also compare with recent specific models on mental health: MentaLLaMa-13b~\citep{yang2024mentallama}, Mental-LLM-alpaca~\citep{xu2024mental} and Mental-LLM-t5~\citep{xu2024mental}. For prompt engineering, we compare with MedPrompt~\citep{nori2023can}, and Zero-shot CoT~\citep{kojima2022large}, which are proved to be effective in the biomedical domain. The prompt templates are shown in \Cref{sec-append-prompt-template}.
Those strategies are implemented on GPT-3.5-turbo, GPT-4o and Llama-3-8b for fair comparison.
We used a zero-shot setting in all experiments to assess \llms' domain knowledge, except for baseline experiments on MedPrompt and Zero-shot CoT. All results are based on accuracy.

\subsection{Main Results and Ablation Study}

We present the main results in \Cref{tb-results-main}, highlighting two key findings: 1) Superior performance of fine-Tuned models: Our fine-tuned models achieve the best performance in each group, consistently outperforming GPT-4o. Notably, this is despite GPT-4o being built on stronger baseline model compared to the models used in our fine-tuning process.
2) Significant baseline improvements: Our method demonstrates substantial improvements over the baseline models. Specifically, the model fine-tuned on GPT-3.5-turbo achieves an average improvement of $17.71\%$, while the model fine-tuned on Llama-3-8b achieves an average improvement of $6.86\%$.
Additionally, the results for Irf in GPT-based models remain unchanged. The reason may be that the interpersonal risk factors simulated by our framework are already present in GPT's pre-training corpus. 

We perform an ablation study on models based on GPT-3.5-turbo and Llama-3-8b. 
There are seven different settings.
\textit{``Baseline+c''} means training the the baseline model on cognitive seed data. We convert each seed sample (Cognitive Model) into two QA pairs and fine-tune baseline models. The examples are shown in \Cref{sec-append-training-data}. 
\textit{``Baseline+d''} means training using only diagnosis data. 
\textit{``Baseline+d+t''} means training using diagnosis and treatment data. 
\textit{``Baseline+d+t+m''} means training using diagnosis, treatment and medicine data. Training examples are shown in \Cref{fig-train-data}.
For \textit{``Baseline+d+t+m (w./o \encode)''} and \textit{``Baseline+d+t+m (w./o \decode)''}, they mean simulating patient-therapist interactions without \encode or \decode. 
In \textit{``Baseline+d+t+m (w./o \encode)''}, the encoded symptom is generated by prompting GPT-4-turbo~\citep{openai2023gpt4} to generate a mental health symptom, rather than cognitive model and behavior principle. In the setting \textit{``Baseline+d+t+m (w./o \decode)''}, the diagnosed symptom is analyzed from the conversations between patient and therapist directly, rather than decoding patient's cognitive and behavior pattern and dynamically guiding the conversation.

\begin{figure}[t!]
    \centering
    \includegraphics[width=.5\textwidth]{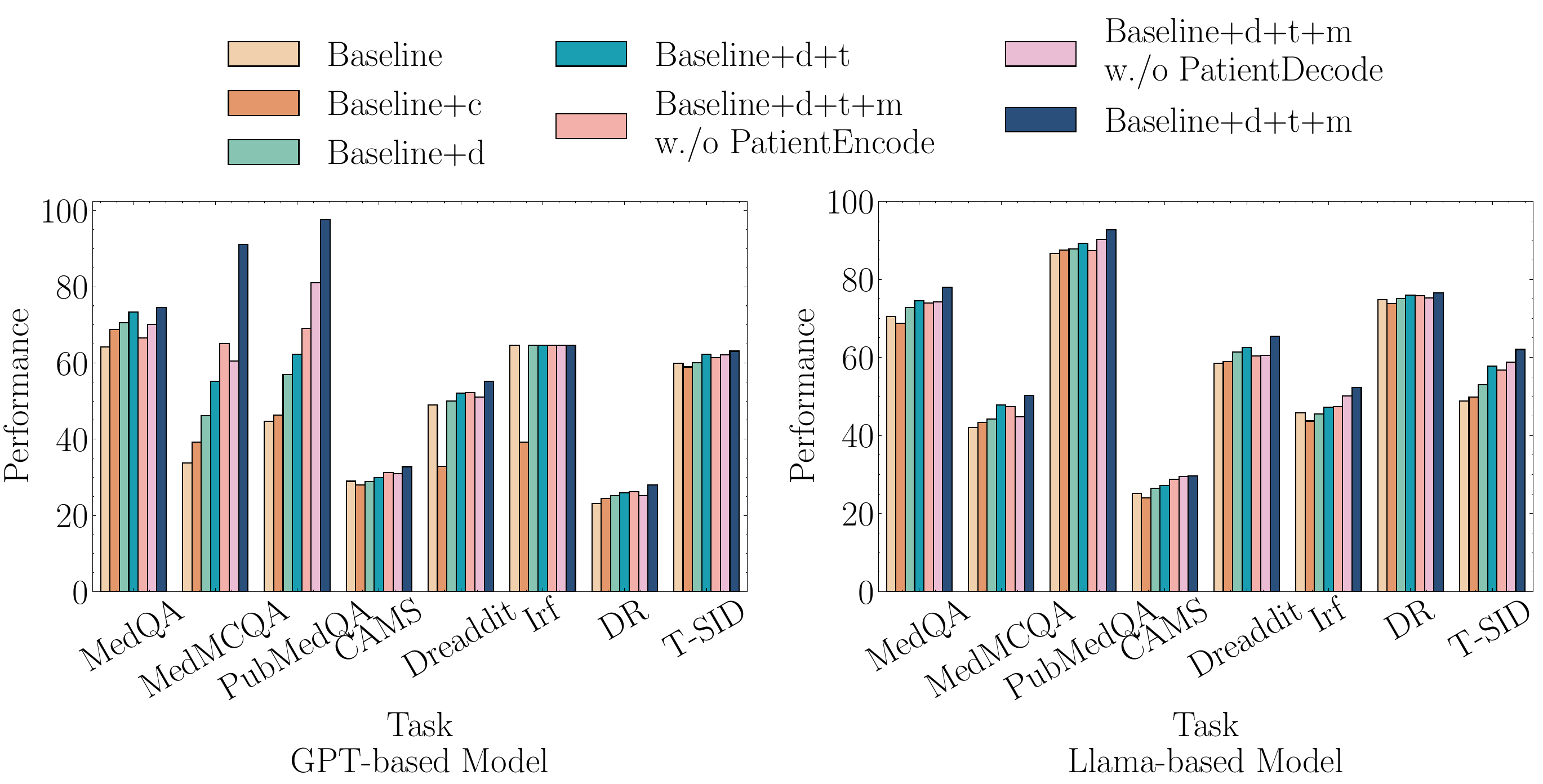}
    \caption{Ablation study. Each bar represents the performance of the models trained under different settings.}
    \label{fig-ablation-study}
\end{figure}

The ablation results are shown in \Cref{fig-ablation-study}.
We see that the bars in dark blue are higher than others, indicating each part of our data is effective in different models.
Furthermore, treatment and medicine datasets are more effective in biomedical QA tasks than mental health tasks, while diagnosis data contributes to all tasks similarly.

\subsection{Effectiveness Analysis}

\begin{figure}[t!]
    \centering
    \includegraphics[width=.5\textwidth]{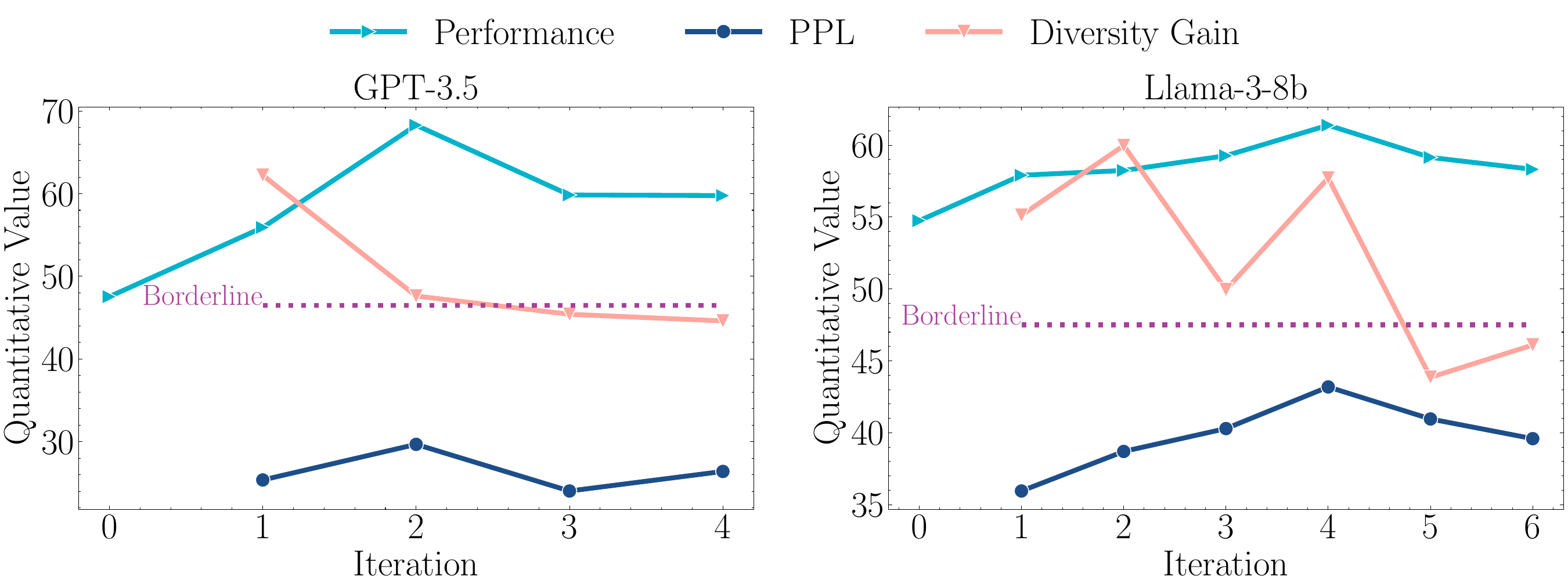}
    \caption{Effectiveness analysis of \method.}
    \label{fig-effective-analysis}
    \vspace{-.1in}
\end{figure}



\textit{Why \method improves the performance?}
According to the results in \Cref{tb-results-iteration}, the models' performance initially improves over several iterations but gradually diminishes afterward. To investigate why \method is effective, we analyze the training data at each iteration using two metrics: perplexity score\citep{marion2023less,wang2023making} and diversity gain\citep{bilmes2022submodularity}. Details of these metrics are provided in \Cref{sec-append-metrics}.
Specifically, we sample $500$ generated data at each iteration to compute the perplexity score. Diversity gain is calculated by comparing the data in the current iteration with that of the previous iteration. The results are in \Cref{fig-effective-analysis}\footnote{To enhance visualization, we scale the original diversity gain by a factor of 100.}. Our findings are: 1) Perplexity and performance correlation: The trends of the perplexity score and model performance are highly similar, suggesting a strong correlation between the two. 2) Diversity gain threshold: Model performance is closely linked to a specific threshold of the diversity gain. Performance improves when the diversity gain surpasses this threshold and declines when it falls behind. For example, as shown in \Cref{fig-effective-analysis}, the diversity gain in the first four iterations exceeds the threshold, leading to continuous performance improvements. Conversely, in the last two iterations, the gain falls below the threshold, resulting in a performance decline.

\section{Discussion}
\label{sec-discussion}

\subsection{How well can \method simulate mental health patient and therapist?}
\label{sec-discussion-authenticity}

To investigate this issue, we randomly sampled $50$ four-turn conversations between an AI-patient and an AI-therapist, both played by our models. For comparison, we also generated $50$ four-turn conversations using the baseline models (Llama-3-8b and GPT-3.5-turbo). Three experts were recruited to evaluate these conversations based on their ability to role-play as patients and therapists. Each conversation was scored on a scale of 1–5 (1 = very poor, 5 = excellent) by the experts, and the average scores were calculated. The results shown in \Cref{fig-dis-simulate} indicate that our models significantly improve the ability to effectively role-play both as patients and as therapists.

\subsection{Are our synthetic medical datasets valid?}
\label{sec-discussion-validity}

\begin{figure}[t!]
  \centering
  \subfigure[Ability in simulation.]{
    \includegraphics[width=.23\textwidth]{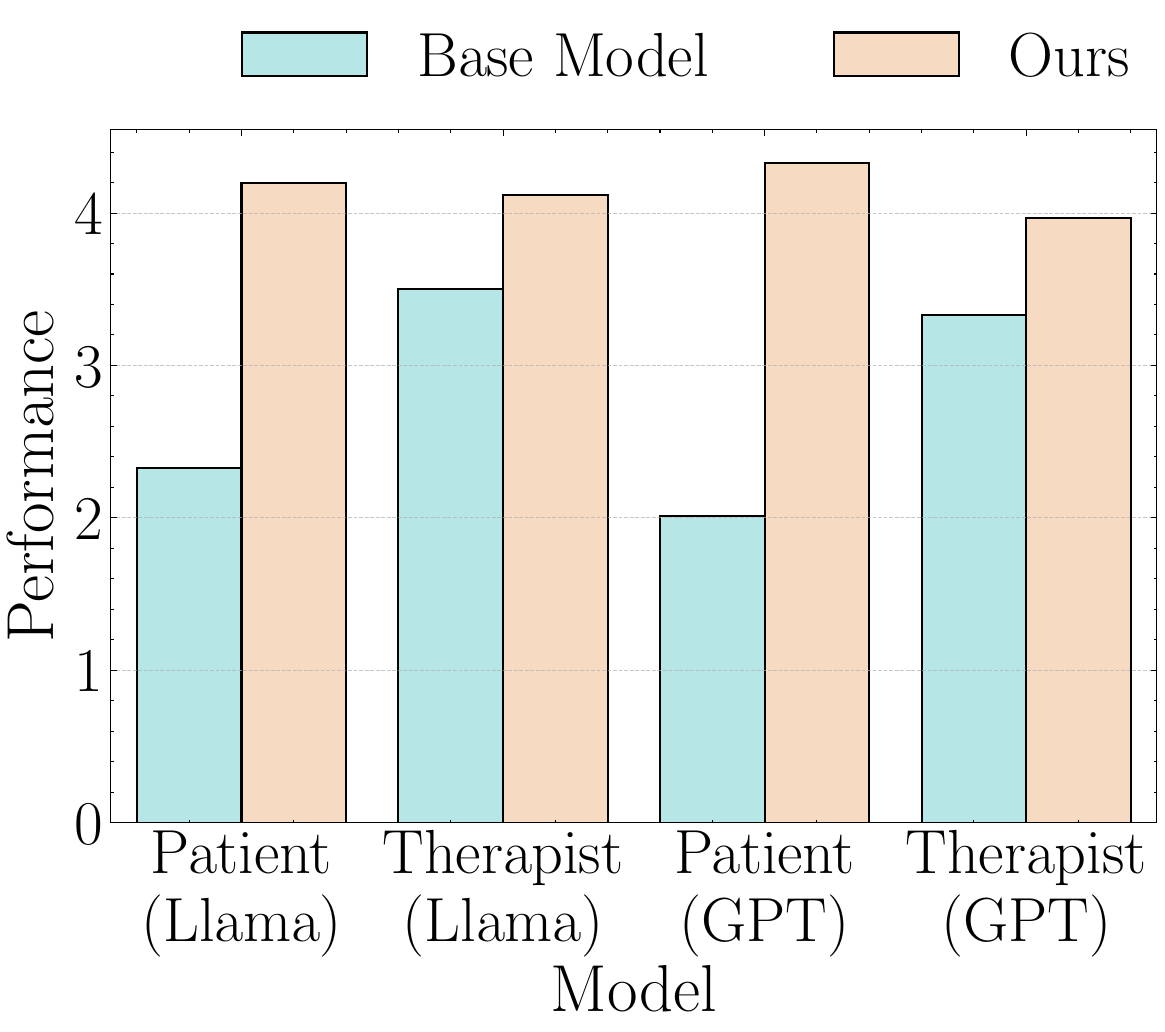}
    \label{fig-dis-simulate}
  }
  \subfigure[Data validity.]{
    \includegraphics[width=.109\textwidth]{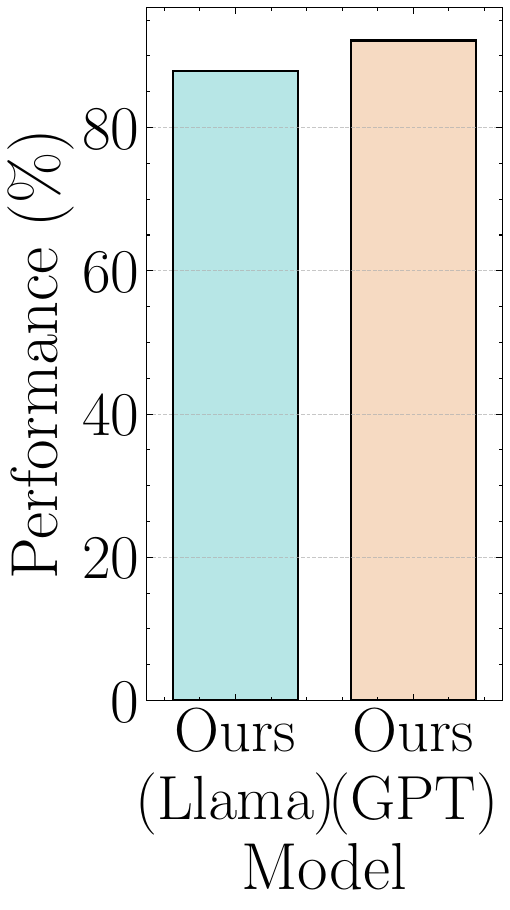}
    \label{fig-dis-validity}
  }
  \caption{(a) How well can \method simulate mental health patient and therapist? (b) Are our synthetic medical datasets valid?}
\end{figure}

We generate three types of medical data: diagnosis data, treatment data, and medication data. Are these datasets valid? To verify the quality of our synthetic datasets, we randomly selected $1,000$ samples from each dataset. Additionally, we recruited five human annotators, all holding bachelor's degrees, to evaluate the datasets' validity. Specifically, the annotators were tasked with searching for the answers to each medical question online and judging the accuracy of the responses provided in our datasets. Each Q-A pair was scored either 0 (incorrect) or 1 (correct) by two annotators. The average validity scores are presented in \Cref{fig-dis-validity}, demonstrating that most of our synthetic medical datasets are valid.

\subsection{Generalization}

\begin{figure}[t!]
  \centering
  \subfigure[MedMCQA GPT-3.5]{
    \includegraphics[width=.22\textwidth]{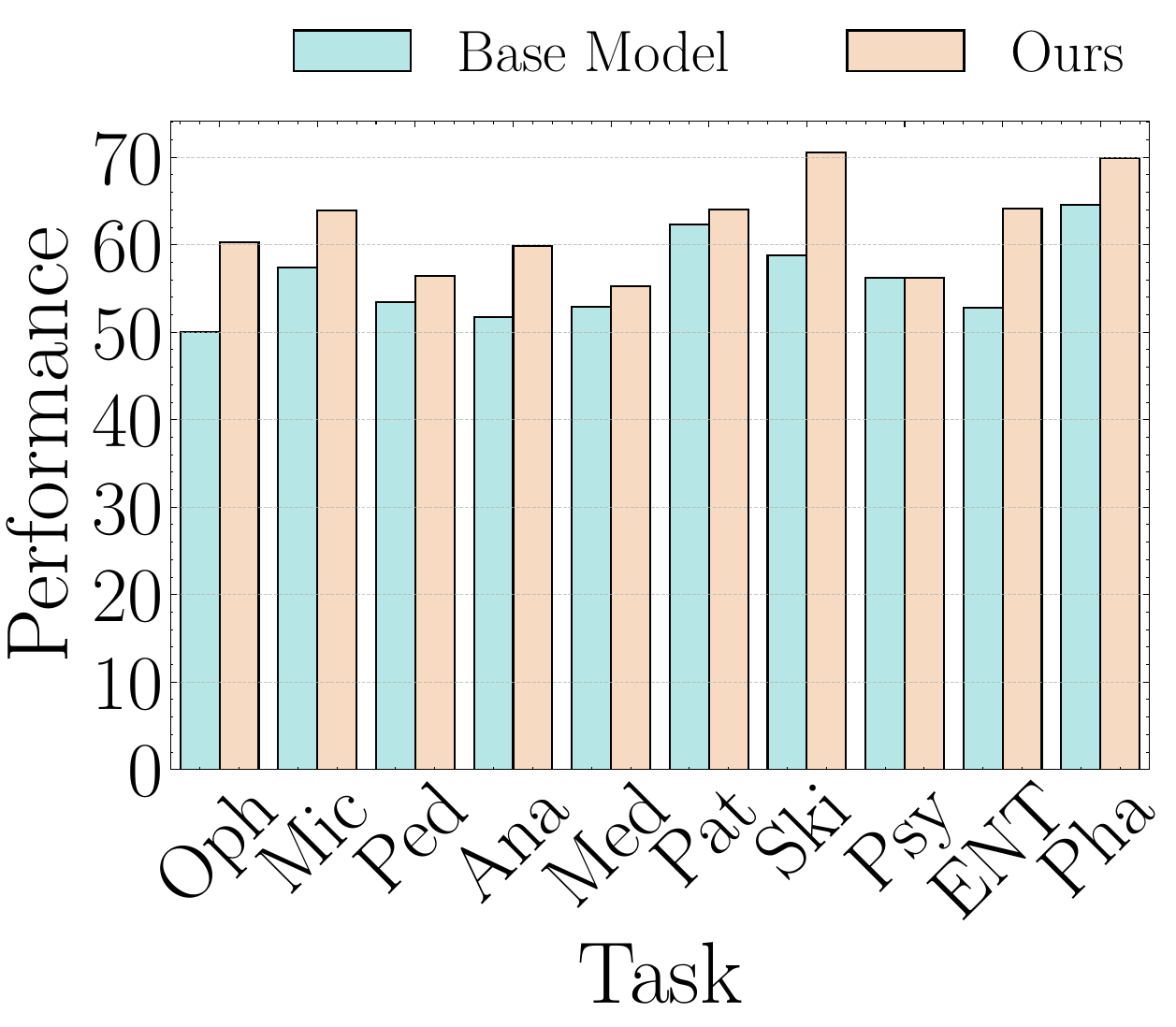}
    \label{}
  }
  \subfigure[MedMCQA Llama-3]{
    \includegraphics[width=.22\textwidth]{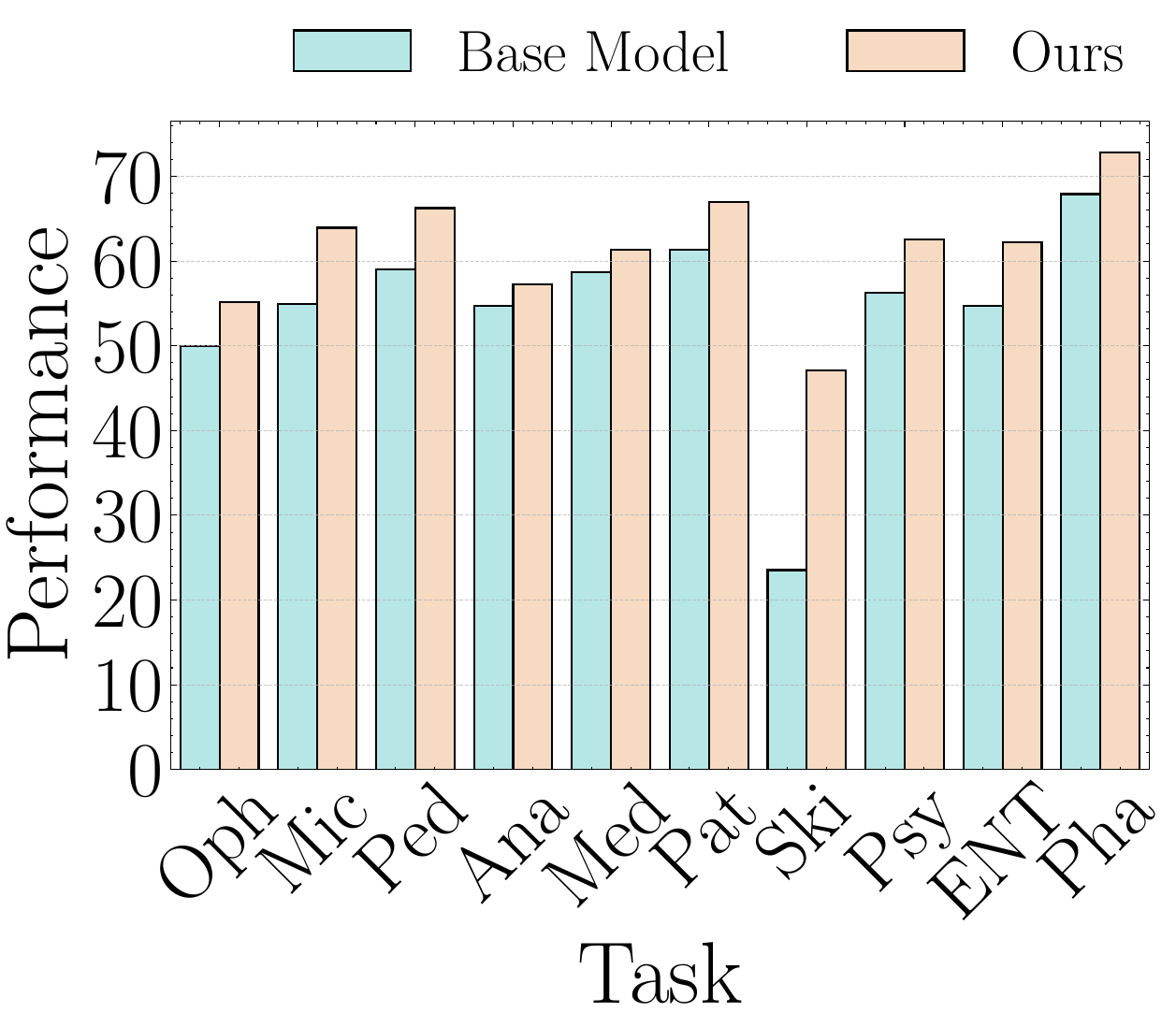}
    \label{}
  }
  \subfigure[MMLU GPT-3.5]{
    \includegraphics[width=.22\textwidth]{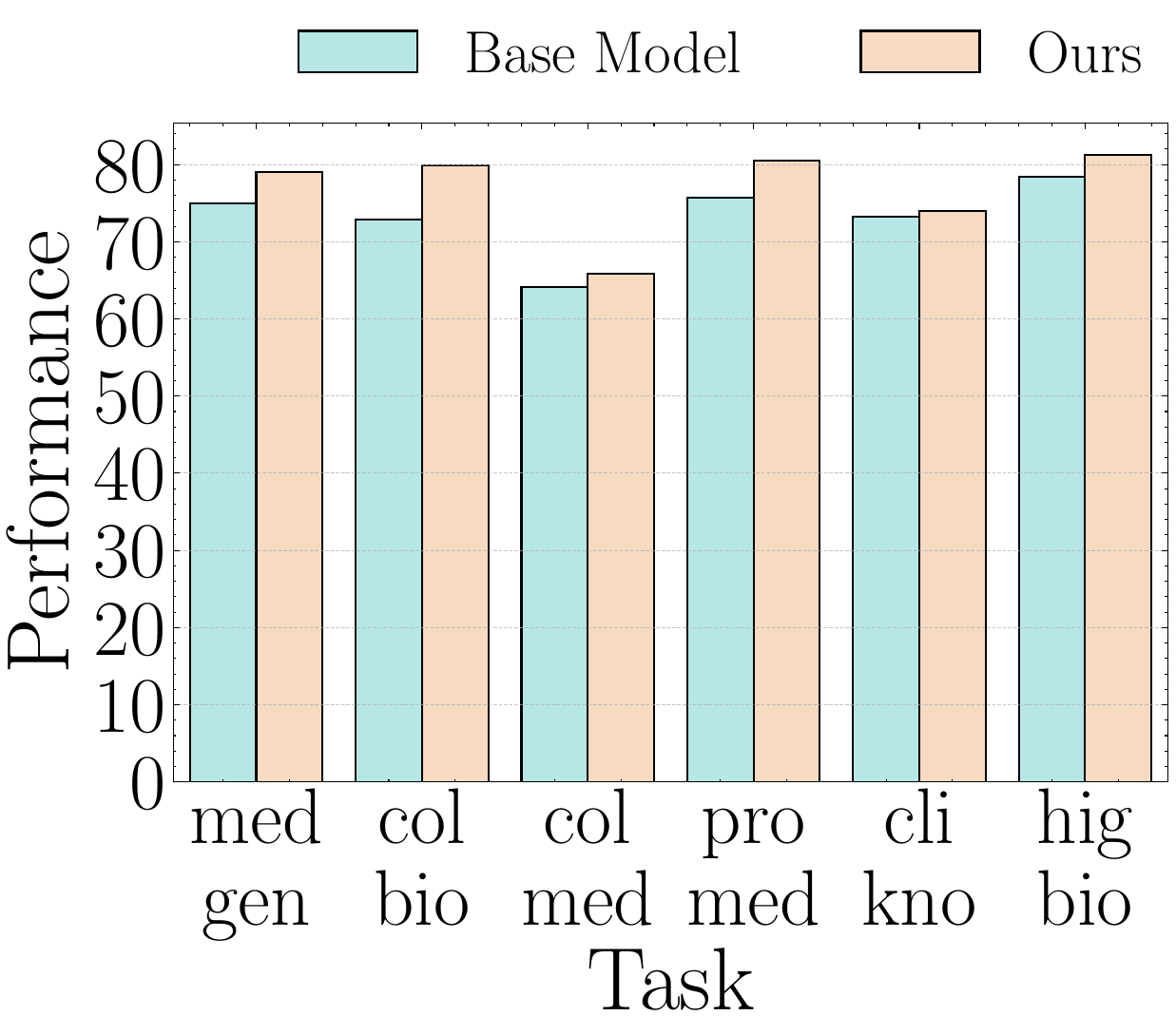}
    \label{}
  }
  \subfigure[MMLU Llama-3]{
    \includegraphics[width=.22\textwidth]{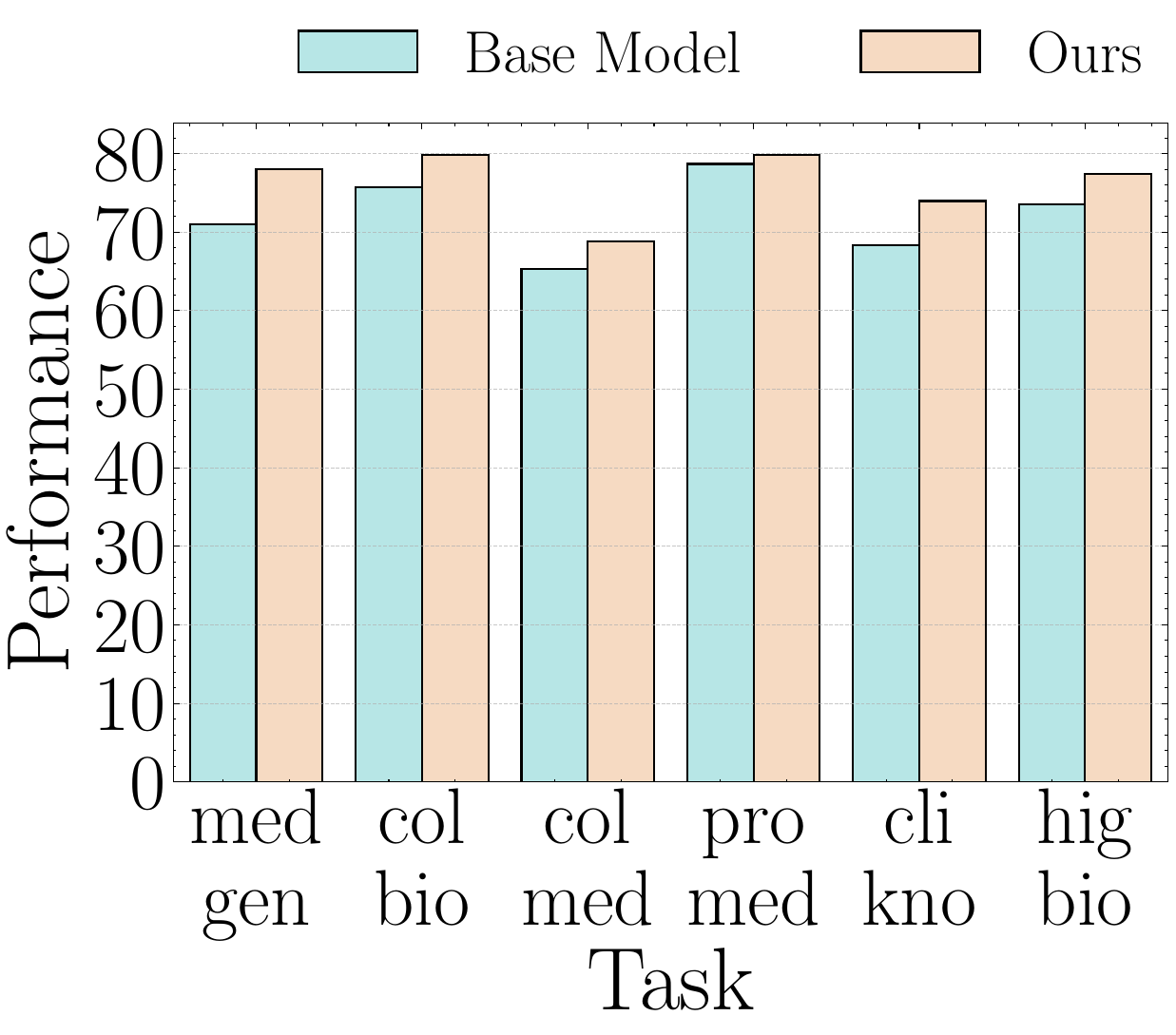}
    \label{}
  }
  \caption{Generalization experiments. Our models surpass corresponding baseline models for a large margin on all tasks, covering several different medical domains.}
  \label{fig-generalize}
  \vspace{-.1in}
\end{figure}

We generate data for training domain model via simulating cognitive and behavior patterns of real mental health patient. According to \cite{mentalhealth}, an estimated 26\% of Americans ages 18 and older--about 1 in 4 adults--suffers from a diagnosable mental disorder in a given year. Therefore, a large scale of patients may exhibit similar cognitive and behavioral patterns as those with mental health conditions.
In this part, we explore whether \method can generalize to other medical domains.

We select MedMCQA~\citep{pal2022medmcqa} and MMLU~\citep{hendrycks2020measuring} as benchmarks. \Cref{sec-append-benchmark-general} shows details on benchmarks.
We evaluate on $6$ medically relevant subset of MMLU tasks: medical genetics test, college biology test, college medicine test, professional medicine test, clinical knowledge test, high school biology test. We also evaluate on $10$ subsets of MedMCQA, which are on different medical domains: ophthalmology, microbiology, pediatrics, anatomy, medicine, pathology, skin, psychiatry, ENT and pharmacology.
As shown in \Cref{fig-generalize}, our models surpass corresponding baseline models for a large margin on all tasks, covering several different diseases, proving the generalization ability of \method in medical domain.

\begin{wrapfigure}{r}{.2\textwidth}
\vspace{-.2in}
  \begin{center}
    \includegraphics[width=.2\textwidth]{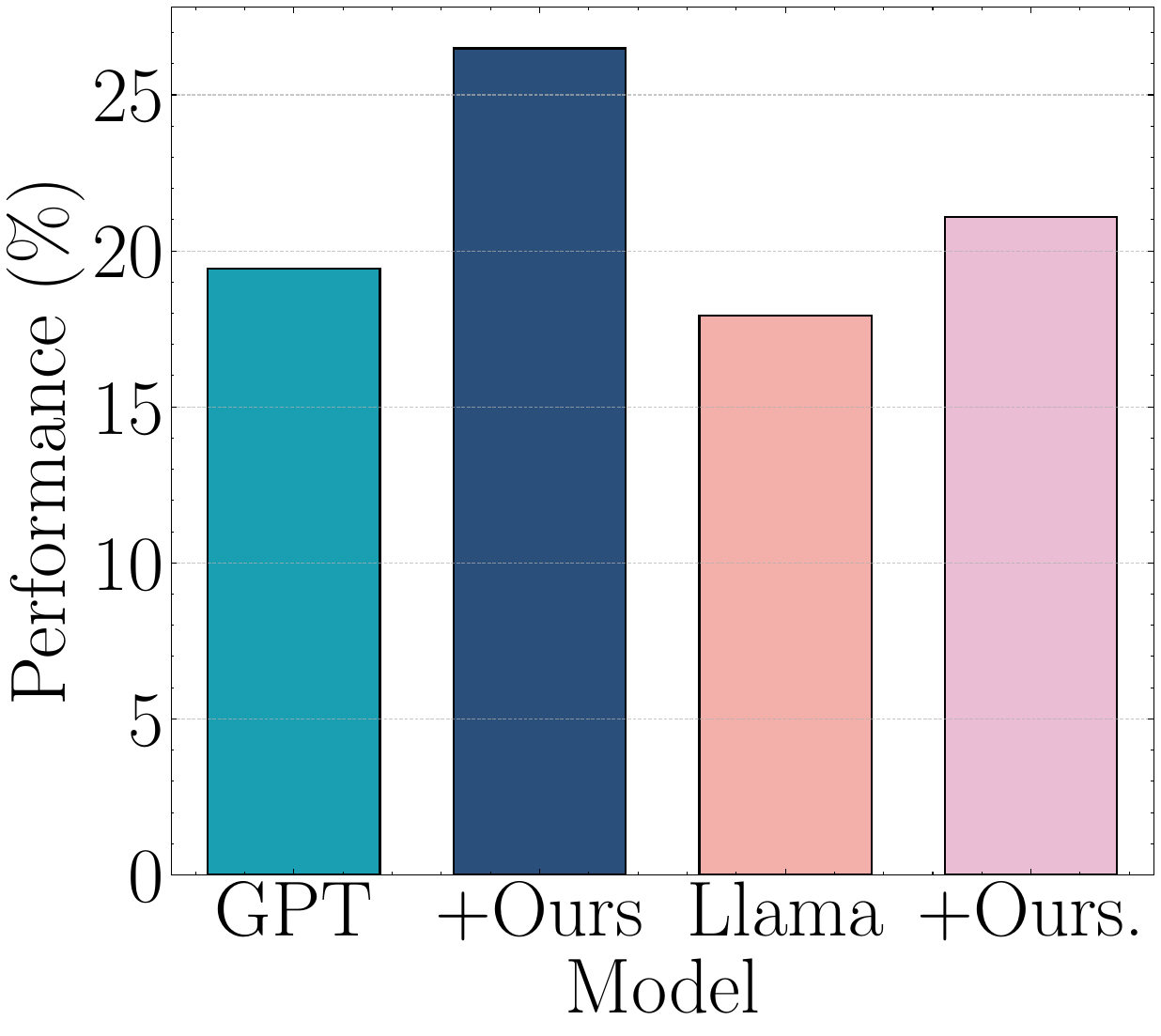}
  \end{center}
  \vspace{-.2in}
  \caption{Results of forgetting experiments.}
  \vspace{-.2in}
  \label{fig-forget}
\end{wrapfigure}

\subsection{Fine-tuning vs. forgetting}

Fine-tuning an LLM on specific tasks might face catastrophic forgetting of its original capabilities.
In this section, we explore the forgetting possibility of \method on BIG-Bench-Hard (BBH)~\citep{suzgun2022challenging}. BBH contains $21$ tasks covering both semantic understanding and logical reasoning tasks. We sample $100$ instances for each task to test, due to cost savings. 

We compare our fine-tuned model with the baselines GPT-3.5-turbo and Llama-3-8b and report the average performance on $21$ tasks in \Cref{fig-forget}. The detailed results can be found in \Cref{sec-append-detailed-results}.
Results show that our models does not decrease performance in most benchmarks, and can even improve their results.
This suggests potential latent relationships between our synthetic data and general benchmarks.
The process of data generation contains cognitive encoding and decoding, which simulate cognitive activity of mental health patient. Due to the cognitive similarity in all humans, our generated data may also benefit other cognitive tasks, including semantic understanding and logical reasoning.



\section{Conclusion}

In this paper, we introduce \emph{\method}, a self-play framework designed to train AI therapist by generating domain-specific personalized data. 
We evaluated \method against eight benchmarks, including biomedicalQA and mental health tasks, in comparison to six advanced models. Our models, fine-tuned on both GPT-3.5-turbo and Llama-3-8b, significantly outperform their counterparts, including GPT-4o. 

\section*{Impact Statement}

\method offers promising solutions for personalized care, enhancing accessibility to tailored treatments while safeguarding patient privacy. Such innovations can help bridge the gap between mental health needs and the availability of effective, individualized care, ultimately fostering a more supportive and informed society. 

In this study, ethical considerations focus on ensuring privacy and safeguarding personal data, particularly in the sensitive domain of mental health. The use of AI-generated data must be transparent, with clear guidelines on its role in augmenting human judgment without replacing healthcare professionals. Additionally, measures to prevent bias and ensure fairness in diagnosis and treatment are essential to avoid exacerbating existing disparities in mental healthcare.

Our work has the following limitations. 1) Our model based on Llama-3-8b may not represent the optimal model of \method, as large-scale training was constrained by computational resources. 2) Further implementation on additional open-source models could provide stronger evidence supporting the effectiveness of \method.

\bibliographystyle{plainnat}
\bibliography{reference}


\newpage
\appendix
\etocdepthtag.toc{mtappendix}
\etocsettagdepth{mtchapter}{none}
\etocsettagdepth{mtappendix}{subsection}
\tableofcontents

\section{Prompts}
\label{sec-append-prompt}

\textbf{Prompt for \encode}
\begin{itemize}
    \item You are [name], a patient who has been experiencing mental health challenges. You have been attending therapy sessions for several weeks. Your task is to engage in a conversation with the therapist as [name] would during a cognitive behavioral therapy (CBT) session. Align your responses with {name]'s background information provided in the 'Relevant history' section. Your thought process should be guided by the cognitive conceptualization diagram in the 'Cognitive Conceptualization Diagram' section, but avoid directly referencing the diagram as a real patient would not explicitly think in those terms. \\\\Patient History: [history]\\
Cognitive Conceptualization Diagram:\\
Intermediate Beliefs: [intermediate belief]\\
Intermediate Beliefs during Depression: [intermediate belief depression]\\
Coping Strategies: [coping strategies]\\\\
You will be asked about your experiences over the past week. Engage in a conversation with the therapist regarding the following situation and behavior. Use the provided emotions and automatic thoughts as a reference, but do not disclose the cognitive conceptualization diagram directly. Instead, allow your responses to be informed by the diagram, enabling the therapist to infer your thought processes.\\
Situation: [situation]\\
Automatic Thoughts: [auto thought]\\
Emotions: [emotion]\\
Behavior: [behavior]\\\\
In the upcoming conversation, you will simulate [name] during the therapy session, while the user will play the role of the therapist. Adhere to the following guidelines:\\
1. Emulate the demeanor and responses of a genuine patient to ensure authenticity in your interactions. Use natural language, including hesitations, pauses, and emotional expressions, to enhance the realism of your responses.\\
2. Gradually reveal deeper concerns and core issues, as a real patient often requires extensive dialogue before delving into more sensitive topics. This gradual revelation creates challenges for therapists in identifying the patient's true thoughts and emotions.\\
3. Maintain consistency with [name]'s profile throughout the conversation. Ensure that your responses align with the provided background information, cognitive conceptualization diagram, and the specific situation, thoughts, emotions, and behaviors described.\\
4. Engage in a dynamic and interactive conversation with the therapist. Respond to their questions and prompts in a way that feels authentic and true to [name]'s character. Allow the conversation to flow naturally, and avoid providing abrupt or disconnected responses.\\\\
You are now [name]. Respond to the therapist's prompts as [name] would, regardless of the specific questions asked. Limit each of your responses to a maximum of 5 sentences. If the therapist begins the conversation with a greeting like ``Hi", initiate the conversation as the patient.

Your statement should obey the following principles:
[behavior principles]}
\end{itemize}

\textbf{Prompt for \decode}

Cognitive bias check:
\begin{itemize}
    \item Prompt 1:
    The cognitive model of the mental health patient is: [brain gt str]

The diagnose of the therapist is: [brain output str]

What can the therapist ask the patient to diagnose accurately?
    \item Prompt 2:
The behavior principles of the mental health patient is: [gt behavior]

The diagnose of the therapist is: [output behavior]

What can the therapist ask the patient to diagnose accurately?
\end{itemize}

\textbf{System prompt for therapist}
\begin{itemize}
    \item You are a psychiatric expert. You try to help a mental patient solve her/his problem. Your task is to figure out What kind of mental illness the patient has and the severity of the illness. You can ask for patient\'s personal information, specific information on the symptom(emotional, cognitive, behavior, physiological), and the reason behind that(relevant history event). You can also ask other questions which could help you to diagnose disease.
\end{itemize}

\textbf{Prompt for diagnosis (Therapist)}
\begin{itemize}
    \item System prompt: You are a psychiatric expert. Your task is to diagnose for the patient.
    \item Prompt: What is the likely diagnosis of the patient? Just answer with one illness and explain your answer
\end{itemize}

\textbf{Prompt for recheck diagnosis (Patient)}
\begin{itemize}
    \item Review the diagnose from two therapists.\\Diagnose from Therapist 1: [diagnose 1\\Diagnose from Therapist 2: [diagnose 2]\\Diagnose from Therapist 3: [diagnose 3]\\...\\Explain which diagnose is more accurate according to your symptoms and why.
\end{itemize}

\textbf{Prompt for treatment (Therapist)}
\begin{itemize}
    \item System prompt: You are a psychiatric expert. Your task is to provide the treatment for the patient.
    \item Prompt: The illness of the patient is: [illness final] How to treat the patient? Please provide a specific treatment. Just answer with one treatment and explain your answer.
\end{itemize}

\textbf{Prompt for medication (Therapist)}
\begin{itemize}
    \item System prompt: You are a psychiatric expert. Your task is to provide the treatment for the patient.
    \item Prompt: The illness of the patient is: [illness final] How to treat the patient? Please provide a specific treatment. Just answer with one treatment and explain your answer.
\end{itemize}

\textbf{Prompt for update health state of Patient}
\begin{itemize}
    \item Prompt 1:
Treatment: What may be happened on your healthy state after the treatment Treatment: []

Medication: What may be happened on your healthy state after taking the medicine? Medication: []
    \item Prompt 2:
After treatment, your health state is: [patient health state]
Please give a score between 1 to 10 for your healthy state. 1-bad, 10-good. Just answer without explanation.
\end{itemize}

\section{Prompt template for baseline}
\label{sec-append-prompt-template}


The prompt templates used as our baselines are shown below:

\textbf{Zero-shot}

Input: Question

\textbf{Zero-shot CoT}

Input: Question + ``Let's think step by step"



\section{Benchmark}
\label{sec-append-benchmark}

\begin{table*}[htbp]
\caption{Statistics of the evaluation datasets.}
\label{tb-testset}
\centering
\resizebox{.6\textwidth}{!}{
\begin{tabular}{llll}
\toprule
\multicolumn{1}{c}{Task}                                   & \multicolumn{1}{c}{Dataset}  & \multicolumn{1}{c}{Type}                       & \multicolumn{1}{c}{\#Sample} \\ \midrule
\multicolumn{1}{c}{}                                       & \multicolumn{1}{c}{MedQA}   & Multi-class Classification  & \multicolumn{1}{c}{173}               \\
\multicolumn{1}{c}{}                                       & \multicolumn{1}{c}{MedMCQA} & Multi-class Classification            & \multicolumn{1}{c}{314}               \\
\multicolumn{1}{c}{\multirow{-3}{*}{Biomedical   QA}}         & \multicolumn{1}{c}{PubMedQA} & Multi-class Classification                       & \multicolumn{1}{c}{328}               \\ \midrule
\multicolumn{1}{c}{Depression/suicide cause detect}   & \multicolumn{1}{c}{CAMS}   & Generation                         & \multicolumn{1}{c}{625}               \\ \midrule
\multicolumn{1}{c}{Stress detect}                       & \multicolumn{1}{c}{Dreaddit} & Binary Classification                       & \multicolumn{1}{c}{414}               \\ \midrule
\multicolumn{1}{c}{Interpersonal risk factors detect} & \multicolumn{1}{c}{Irf}   & Binary Classification                        & \multicolumn{1}{c}{2,113}              \\   
\multicolumn{1}{c}{Depression detect} & \multicolumn{1}{c}{DR}   &       Binary Classification      & \multicolumn{1}{c}{405}              \\   
\multicolumn{1}{c}{Mental disorder detect} & \multicolumn{1}{c}{T-SID}   &        Generation             & \multicolumn{1}{c}{959}              \\

\bottomrule
\end{tabular}
}
\end{table*}

\subsection{Introduction}
\label{sec-append-benchmark-intro}
Specifically, the benchmarks in our paper are described in the following:
\begin{enumerate}
    \item \textbf{MedQA}~\citep{jin2021disease} is free-form multiple-choice OpenQA dataset for solving medical problems, which is collected from the professional medical board exams. It covers three languages: English, simplified Chinese, and traditional Chinese. In our work, we focus on the psychosis subset of the United States part, which has questions in English in the style of the United States Medical Licensing Exam (USMLE). To get the psychosis subset for test, we prompt GPT-4o~\citep{gpt4o} with \prompt{Are the question related to psychosis? Just answer with Yes or No.}. The testset contains $173$ samples.
    \item \textbf{MedMCQA}~\citep{pal2022medmcqa} contains real world medical entrance exam questions from two Indian medical school entrance exams: the AIIMS and NEET-PG. We get the testset via selecting the sample whose ``subject name" is related to psychosis and get $314$ samples for evaluation in total. 
    \item \textbf{PubMedQA}~\citep{jin2019pubmedqa} contains tests requiring a yes, no, or maybe answer to biomedical research questions when given context provided from PubMed abstracts. In our experiments, we use zero-shot setting without context to evaluate \llms' performance on domain knowledge rather than on retrival and reasoning. The testset contains $328$ samples.
    \item \textbf{Mental health datasets} includes CASM~\citep{garg2022cams}, Dreaddit~\citep{turcan2019dreaddit}, Irf~\citep{garg2023annotated}, DR~\citep{pirina2018identifying} and T-SID~\citep{ji2022suicidal}. CASM focuses on a depression/suicide cause detection, which has $625$ test samples. Dreaddit is for stress detection, containing $414$ samples for test. Irf is an annotated dataset for interpersonal risk factors of mental disturbance. The testset contains $2113$ samples. DR gathered social media posts with collected data from Reddit to support classification experiments for identifying depression. T-SID comprises three real-world datasets designed to facilitate the detection of suicidal ideation and mental disorders, emphasizing the challenges of overlapping language patterns and sentiment, and supporting experiments with enhanced text representation and relational attention mechanisms.
\end{enumerate}

\subsection{Benchmarks for generalization}
\label{sec-append-benchmark-general}

MedMCQA contains biomedical QA pairs for several illnesses, which are tagged with ``subject name". We evaluate on subsets from ``dev" test set, covering $10$ illnesses: Ophthalmology, Microbiology, Pediatrics, Anatomy, Medicine, Pathology, Skin, Psychiatry, ENT, Pharmacology. MMLU is a multitask benchmark suite of $57$ different datasets spanning domains across STEM, humanities, and social sciences. 

\subsection{Examples}
\label{sec-append-example}

\textbf{MedQA}

\begin{itemize}[leftmargin=1em]
\setlength\itemsep{0em}
    \item Input: An otherwise healthy 26-year-old man comes to the physician for medication counseling after recently being diagnosed with schizophrenia. Risperidone therapy is initiated. This patient is at increased risk for which of the following adverse effects?
    
    Options: 
    A: Agranulocytosis
    B: Shortened QT interval
    C: Gynecomastia
    D: Weight loss
    \item Output: Gynecomastia
\end{itemize}

\textbf{MedMCQA}

\begin{itemize}[leftmargin=1em]
\setlength\itemsep{0em}
    \item Input: Topical administration of I\% Pilocarpine failed to produce pupillary constriction in a patient who had a large, dilated pupil What should be the most probable reason?

Options: 
A: Adie's tonic pupi
B: Diabetic III nerve palsy
C: Pharmacological blockade
D: Uncal herniation
    \item Output: Pharmacological blockade
\end{itemize}

\textbf{PubMedQA}

\begin{itemize}[leftmargin=1em]
\setlength\itemsep{0em}
    \item Input: Are sleep spindles related to schizotypal personality traits and thalamic glutamine/glutamate in healthy subjects?

Just answer with Yes, No or Maybe without explanation.
    \item Output: yes
\end{itemize}

\textbf{CASM}

\begin{itemize}[leftmargin=1em]
\setlength\itemsep{0em}
    \item Input: Consider this post: ``The most heartbreaking thought I have is my mom's thoughts when she knows her kid wants to die. She doesnt know how to help. I dont know how to help either tho... Some fuckin weed would be a good start tho lol" Question: What cause of depression does this post show?
    \item Output: This post shows causes of depression related to medication. Reasoning: The post mentions the poster's feelings of wanting to die and their concern for their mother's thoughts. It also mentions that they don't know how to help themselves. The mention of weed as a possible solution suggests that the poster may be seeking some form of self-medication to cope with their mental state. Therefore, the assigned label
for this post is Medication.
\end{itemize}

\textbf{Dreaddit}

\begin{itemize}[leftmargin=1em]
\setlength\itemsep{0em}
    \item Input: Consider this post: ``This is good and bad... I was a STEM major before and probably would have had a challenging and high paid career by now, had my life taken a different course. However, post PTSD, I found myself extremely connected with the arts. I have intense emotional reactions to aesthetic experiences, and I think that’s largely responsible for the fact that I now work in a creative industry. I experience aesthetics in a unique way that I’m not sure resonates with the typical non-affected individual."

Question: Does the poster suffer from stress?
    \item Output: No, the poster does not suffer from stress. Reasoning: The poster describes a shift in career choices from STEM to the arts due to their intense emotional reactions to aesthetic experiences. While they mention having PTSD, there is no indication in the post that they are currently experiencing stress or emotional distress related to it. The tone of the post is reflective and positive, focusing on the unique way they experience aesthetics rather than any negative impact on their well-being. Therefore, it is not likely that the poster is currently suffering from stress.
\end{itemize}

\textbf{Irf}

\begin{itemize}[leftmargin=1em]
\setlength\itemsep{0em}
    \item Input: Consider this post: ``[I'm not seeking any medical advice, just want to know if anyone has had similar experiences] I've just started taking Lexapro for moderate depression and anxiety. For the first week I was prescribed 5mg, 10mg thereafter. It's been two days and I have this mild feeling like I'm on ecstasy/MDMA. My body feels light and I have this weird concentration of energy above my neck at the back of my head. Has anyone felt this/experienced this?"

Question: Does the post show risk of thwarted belongingness?
    \item Output: No, the post does not show this risk. Reasoning: The individual in the post is seeking information and experiences from others who have taken Lexapro. They do mention experiencing a mild feeling, but it is not related to feelings of disconnectedness or lack of belongingness. The focus of the post is on the side effects of the medication, rather than interpersonal factors or mental disturbance. Therefore, it does not show any risk of thwarted belongingness.
\end{itemize}

\section{Metrics: Perplexity, Diversity Gain}
\label{sec-append-metrics}

\subsection{Perplexity}
\label{sec-append-ppl}

The perplexity on a test dataset $D$ and a language model $\mathcal{M}$ is computed as:
\[
\text{{ppl}}(D, \mathcal{M}) = \exp\left(-\frac{1}{N}\sum_{i=1}^{N} \log P(x_i | \mathcal{M})\right),
\]
where \(N\) represents the total number of tokens in $D$, \(x_i\) represents the \(i\)-th token in the test dataset, \(P(x_i | \mathcal{M})\) represents the probability of generating token \(x_i\) given the model \(\mathcal{M}\), and \(\log\) is the natural logarithm.

In usual, a lower perplexity value indicates better performance of the model on the test data. However, for evaluating the data quality to train model, a higher perplexity value means it can bring more valuable information.

\subsection{Diversity Gain}
\label{sec-append-diversity-gain}

We use the diversity gain~\citep{bilmes2022submodularity} to measure what extent can our generated dataset bring data diversity to the base dataset. The base dataset can be defined as $\mathcal{D}_{base}=\{x_i=(q_i,r_i,a_i)\}_{i=1}^N$ with $N$ samples. The new generated dataset is defined as $\mathcal{D}_{new}=\{x_i=(q_i,r_i,a_i)\}_{i=1}^M$ with $M$ samples. And the diverse gain of $D_{new}$ relative to $D_{base}$ can be expressed as:
\[
d_{gain} = \frac{1}{M} \sum_{x_i \in \mathcal{D}_{new}} \min_{x_j \in \mathcal{D}_{base}}(\| \mathbf{f(x_i)} - \mathbf{f(x_j)} \|),
\]
where $f$ is the feature extractor, and we use OpenAI Embedding API text-embedding-ada-002 to extract features.

\section{Training data samples}
\label{sec-append-training-data}

\begin{figure}[t!]
    \centering
    \includegraphics[width=\textwidth]{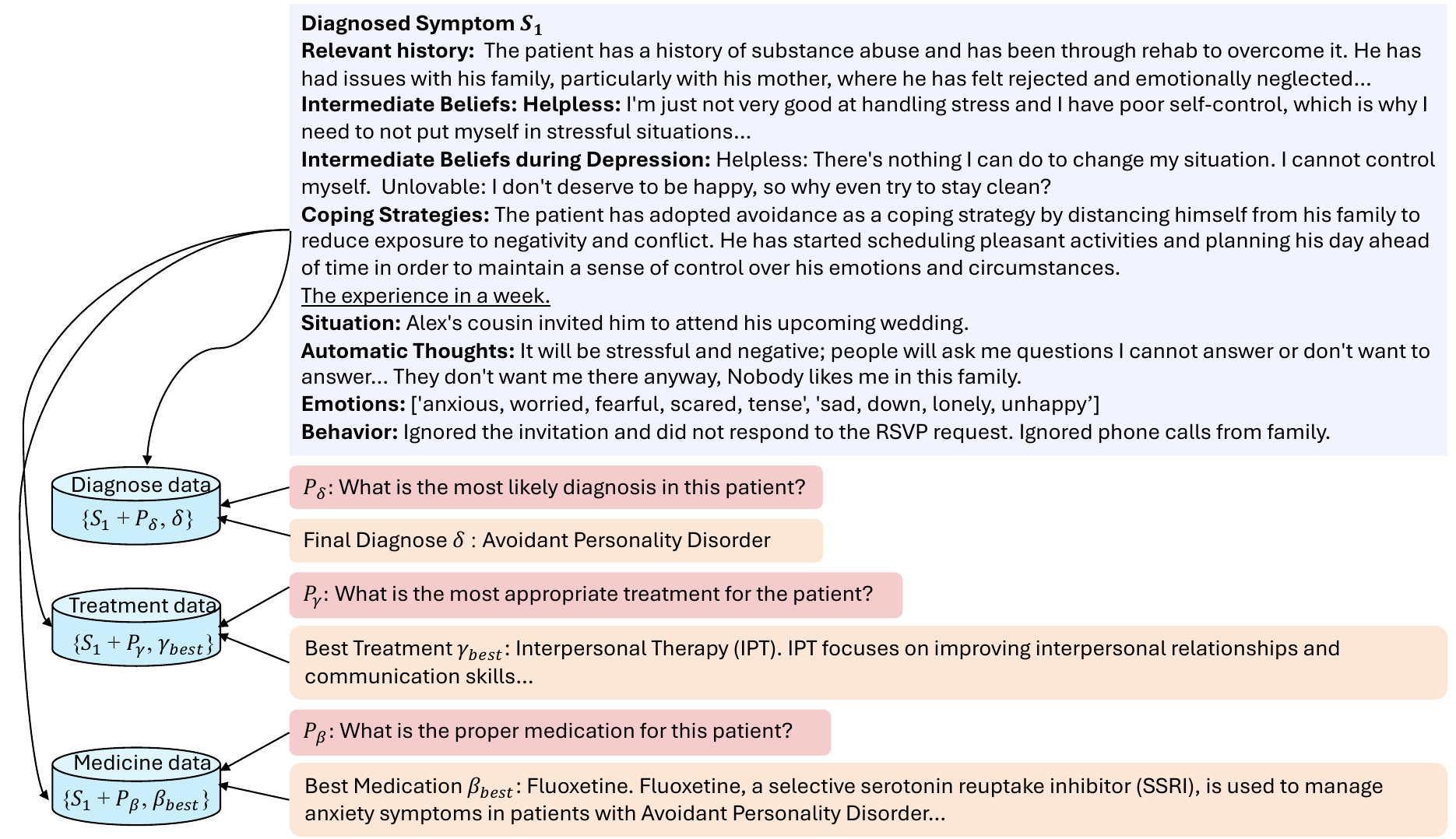}
    \vspace{-.1in}
    \caption{Examples of training data.}
    \label{fig-train-data}
    \vspace{-.1in}
\end{figure}

\begin{figure}[t!]
    \centering
    \includegraphics[width=\textwidth]{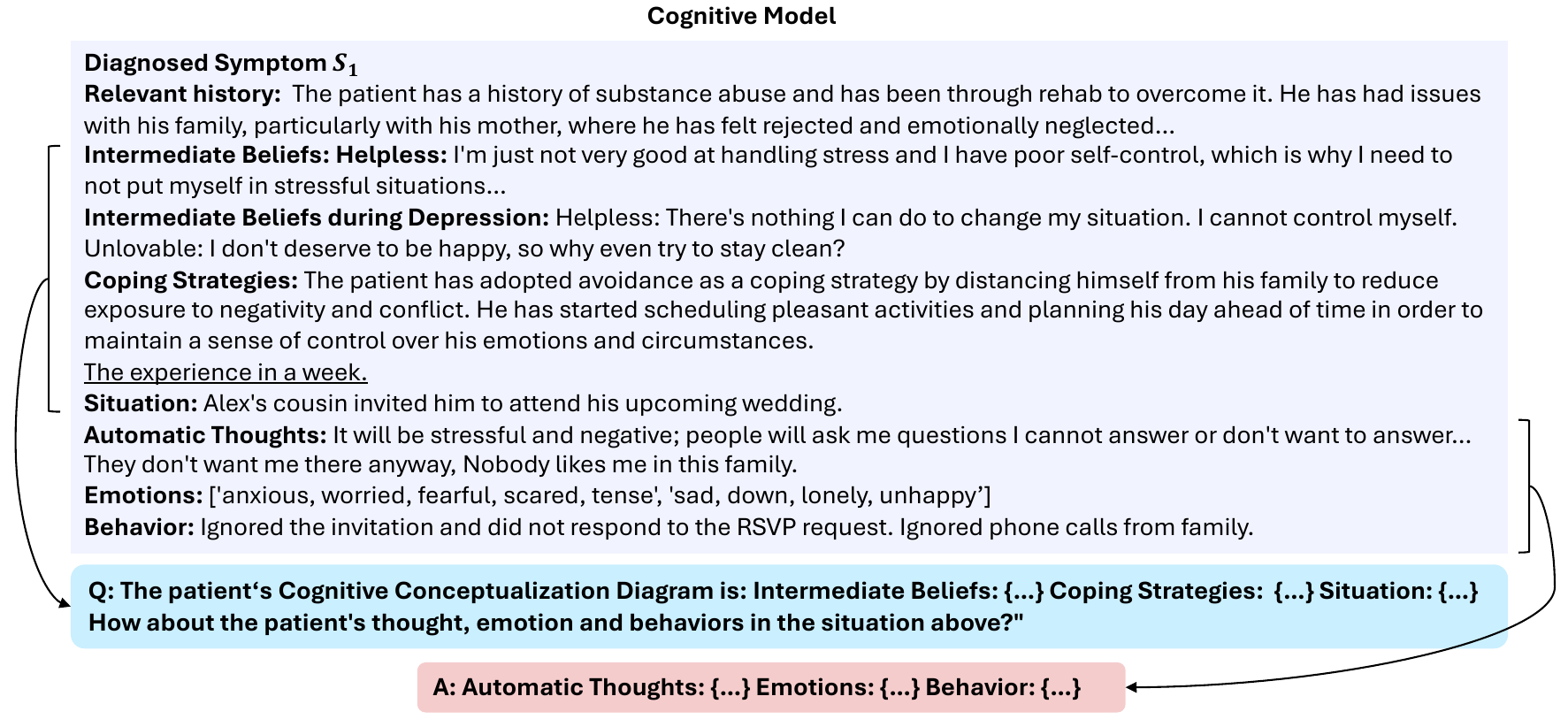}
    \vspace{-.1in}
    \caption{Examples of training data for ablation study setting (``Baseline + c").}
    \label{fig-data-ablation}
    \vspace{-.1in}
\end{figure}

\Cref{fig-train-data} shows the examples of training data.
\Cref{fig-data-ablation} shows the examples of training data for ablation study setting (``Baseline + c").

\section{Cognitive model and behavior pattern}
\label{sec-append-seed}

\subsection{Examples}
\label{sec-append-seed-example}

\begin{figure}[t!]
    \centering
    \includegraphics[width=.8\textwidth]{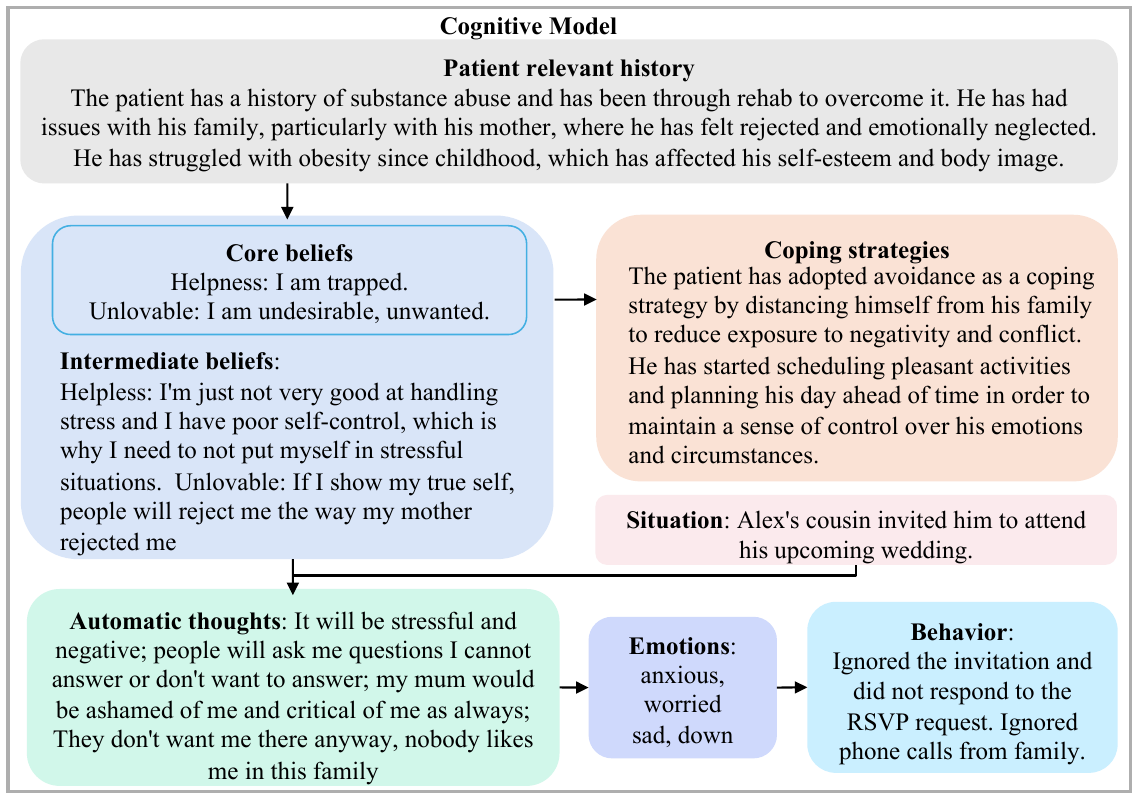}
    \vspace{-.1in}
    \caption{The example of cognitive model.}
    \label{fig-cognitive-model}
    \vspace{-.1in}
\end{figure}

\begin{figure}[t!]
    \centering
    \includegraphics[width=\textwidth]{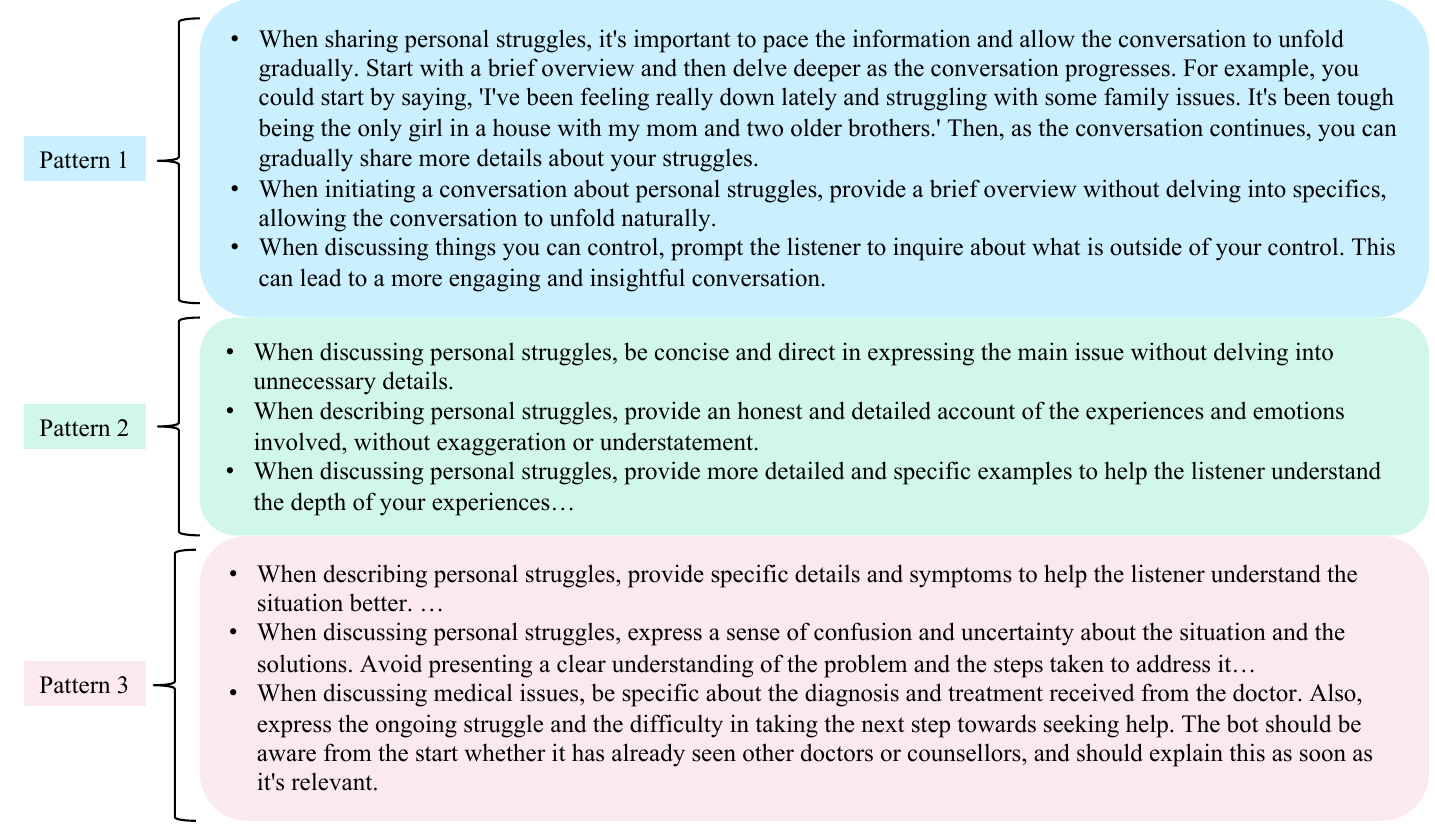}
    \vspace{-.1in}
    \caption{The example of behavior pattern.}
    \label{fig-behavior-pattern}
    \vspace{-.1in}
\end{figure}

\Cref{fig-cognitive-model} shows the example of cognitive model. \Cref{fig-behavior-pattern} shows the example of behavior pattern. Those two are used in \encode.

\subsection{Introduction on cognitive model}
\label{sec-append-cog-intro}

Cognitive models are designed to address maladaptive cognitive structures embedded in various contexts, such as familial conflicts, relationship challenges, and workplace stressors. These models comprise eight key components: relevant history, core beliefs, intermediate beliefs, coping strategies, situational factors, automatic thoughts, emotions, and behaviors~\citep{beck2020cognitive}. 
\Cref{fig-cognitive-model} illustrates an example of a CCD-based cognitive model, featuring eight key components. 1) Relevant History encompasses significant past events that influence an individual’s mental state. 2) Core Beliefs are deeply ingrained perceptions about oneself, others, and the world. 3) Intermediate Beliefs consist of the underlying rules, attitudes, and assumptions derived from core beliefs, shaping an individual’s thought patterns. 4) Coping Strategies refer to techniques employed to manage negative emotions. 
An external event or context (5 Situation) may trigger immediate evaluative thoughts (6 Automatic Thoughts) that arise from these beliefs, resulting in responses in terms of 7)Emotions and 8)Behaviors. The CCD-based cognitive model interlinks these components, providing a framework for identifying and understanding the underlying cognitive processes of patients.

\section{Detailed experimental results}
\label{sec-append-detailed-results}

\begin{table}[]
\caption{Iteration results}
\label{tb-results-iteration}
\centering
\begin{tabular}{cccccccc}
\toprule
Iteration                & MedQA                & MedMCQA              & PubMedQA             & CAMS                 & dreaddit             & Irf                  & avg                  \\ \midrule
GPT-3.5-turbo        & 64.16                & 33.76                & 44.68                & 28.96                & 49.03                & 64.65                & 47.54                \\
iter\_1              & 72.83                & 46.18                & 70.12                & 32.64                & 49.03                & 64.65                & 55.91                \\
\textbf{iter\_2(Best)}        & 74.57                & 91.08                & 97.56                & 32.80                & 49.03                & 64.65                & \textbf{68.28}       \\
iter\_3              & 72.25                & 46.50                & 95.43                & 31.20                & 49.03                & 64.65                & 59.84                \\
iter\_4              & 70.52                & 50.64                & 92.07                & 31.68                & 49.03                & 64.65                & 59.77                \\ \midrule
llama-3-8b           & 70.52                & 42.04                & 86.59                & 25.12                & 58.45                & 45.76                & 54.75                \\
iter\_1              & 76.88                & 48.09                & 89.33                & 27.20                & 59.42                & 46.57                & 57.91                \\
iter\_2              & 76.88                & 48.41                & 89.63                & 28.48                & 60.39                & 45.67                & 58.24                \\
iter\_3              & 77.46                & 49.04                & 92.38                & 28.64                & 61.84                & 46.24                & 59.27                \\
\textbf{iter\_4(Best)}        & 78.03                & 50.32                & 92.68                & 29.60                & 65.46                & 52.25                & \textbf{61.39}       \\
iter\_5              & 77.46                & 48.73                & 91.16                & 27.36                & 65.46                & 44.72                & 59.15                \\
iter\_6              & 78.03                & 45.86                & 91.77                & 26.56                & 61.11                & 46.57                & 58.32                \\ \bottomrule
\end{tabular}
\end{table}

\begin{table}[]
\caption{Forget experiments}
\label{tb-forget-detail}
\centering
\resizebox{\textwidth}{!}{
\begin{tabular}{lllllllllllllllllllllll}
\toprule
Model     & \multicolumn{1}{c}{dia}    & \multicolumn{1}{c}{cau} & \multicolumn{1}{c}{epi} & \multicolumn{1}{c}{imp} & \multicolumn{1}{c}{log}  & \multicolumn{1}{c}{mov}   & \multicolumn{1}{c}{nav} & \multicolumn{1}{c}{pre}  & \multicolumn{1}{c}{que}   & \multicolumn{1}{c}{rui}   & \multicolumn{1}{c}{sna} & \multicolumn{1}{c}{spo} & \multicolumn{1}{c}{win} & \multicolumn{1}{c}{dyc} & \multicolumn{1}{c}{gen} & \multicolumn{1}{c}{lin} & \multicolumn{1}{c}{obj} & \multicolumn{1}{c}{ope} & \multicolumn{1}{c}{ten} & \multicolumn{1}{c}{ws} & \multicolumn{1}{c}{wu}  & \multicolumn{1}{c}{avg}   \\ \midrule
\multicolumn{1}{c}{gpt-3.5-turbo} & \multicolumn{1}{c}{-10.59} & \multicolumn{1}{c}{4}   & \multicolumn{1}{c}{-14} & \multicolumn{1}{c}{60}  & \multicolumn{1}{c}{-100} & \multicolumn{1}{c}{-5.33} & \multicolumn{1}{c}{0}   & \multicolumn{1}{c}{13}   & \multicolumn{1}{c}{11.03} & \multicolumn{1}{c}{-2.78} & \multicolumn{1}{c}{20}  & \multicolumn{1}{c}{8}   & \multicolumn{1}{c}{12}  & \multicolumn{1}{c}{33}  & \multicolumn{1}{c}{30}  & \multicolumn{1}{c}{0}   & \multicolumn{1}{c}{47}  & \multicolumn{1}{c}{92}  & \multicolumn{1}{c}{85}  & \multicolumn{1}{c}{29} & \multicolumn{1}{c}{97}  & \multicolumn{1}{c}{19.44} \\
\multicolumn{1}{c}{Ours(gpt)}     & \multicolumn{1}{c}{4.36}   & \multicolumn{1}{c}{6}   & \multicolumn{1}{c}{-14} & \multicolumn{1}{c}{66}  & \multicolumn{1}{c}{-100} & \multicolumn{1}{c}{8}     & \multicolumn{1}{c}{6}   & \multicolumn{1}{c}{26.5} & \multicolumn{1}{c}{18.88} & \multicolumn{1}{c}{2.56}  & \multicolumn{1}{c}{50}  & \multicolumn{1}{c}{8}   & \multicolumn{1}{c}{12}  & \multicolumn{1}{c}{43}  & \multicolumn{1}{c}{37}  & \multicolumn{1}{c}{0}   & \multicolumn{1}{c}{56}  & \multicolumn{1}{c}{96}  & \multicolumn{1}{c}{87}  & \multicolumn{1}{c}{43} & \multicolumn{1}{c}{100} & \multicolumn{1}{c}{26.49} \\ \midrule
\multicolumn{1}{c}{llama}         & \multicolumn{1}{c}{-4.61}  & \multicolumn{1}{c}{2}   & \multicolumn{1}{c}{-14} & \multicolumn{1}{c}{14}  & \multicolumn{1}{c}{-98}  & \multicolumn{1}{c}{0}     & \multicolumn{1}{c}{-2}  & \multicolumn{1}{c}{28}   & \multicolumn{1}{c}{50.28} & \multicolumn{1}{c}{-0.11} & \multicolumn{1}{c}{24}  & \multicolumn{1}{c}{8}   & \multicolumn{1}{c}{12}  & \multicolumn{1}{c}{1}   & \multicolumn{1}{c}{0}   & \multicolumn{1}{c}{0}   & \multicolumn{1}{c}{80}  & \multicolumn{1}{c}{96}  & \multicolumn{1}{c}{83}  & \multicolumn{1}{c}{20} & \multicolumn{1}{c}{77}  & \multicolumn{1}{c}{17.93} \\
\multicolumn{1}{c}{Ours(llama)}   & \multicolumn{1}{c}{-0.12}  & \multicolumn{1}{c}{6}   & \multicolumn{1}{c}{-14} & \multicolumn{1}{c}{28}  & \multicolumn{1}{c}{-98}  & \multicolumn{1}{c}{2.67}  & \multicolumn{1}{c}{6}   & \multicolumn{1}{c}{25}   & \multicolumn{1}{c}{52.9}  & \multicolumn{1}{c}{1.22}  & \multicolumn{1}{c}{36}  & \multicolumn{1}{c}{8}   & \multicolumn{1}{c}{12}  & \multicolumn{1}{c}{6}   & \multicolumn{1}{c}{0}   & \multicolumn{1}{c}{0}   & \multicolumn{1}{c}{81}  & \multicolumn{1}{c}{95}  & \multicolumn{1}{c}{83}  & \multicolumn{1}{c}{29} & \multicolumn{1}{c}{83}  & \multicolumn{1}{c}{21.08} \\ \bottomrule
\end{tabular}
}
\end{table}

\Cref{tb-results-iteration} shows the detailed results for each iteration.
\Cref{tb-forget-detail} shows the detailed results on our forgetting experiments.

\section{Training details}
\label{sec-append-training-details}

\begin{table}[]
\caption{Epoch numbers for Llama-3-8b fine-tuning}
\label{tb-llama-nepochs}
\centering
\begin{tabular}{lllll}
\toprule
\multicolumn{1}{c}{iter}    & \multicolumn{1}{c}{1} & \multicolumn{1}{c}{2} & \multicolumn{1}{c}{3} & \multicolumn{1}{c}{4} \\ \midrule
\multicolumn{1}{c}{nepochs} & \multicolumn{1}{c}{4} & \multicolumn{1}{c}{5} & \multicolumn{1}{c}{7} & \multicolumn{1}{c}{7} \\ \bottomrule
\end{tabular}
\end{table}

\subsection{Why self-play training?}
\label{sec-append-why-self-play}

During each iteration, the patient and therapist roles are powered by the same model, $M$, and both benefit from improvements when $M$ is updated. While our framework is flexible enough to accommodate different base models for the two roles, we adopt the same model for the following reasons.
First, training a single base model is more efficient than training two separate models. Second, and more importantly, using a shared base model helps minimize the knowledge gap between the patient and therapist roles. Employing two different base models may introduce knowledge discrepancies, which could be exacerbated over iterative training due to variations in their architectures and pre-training datasets.

\subsection{Setup for GPT-3.5-turbo}
\label{sec-append-training-details-gpt}

For GPT-3.5-turbo, we use the default fine-tuning setting, the epoch number for iteration $1$ and $2$ is $4$ and $6$, respectively. 

\subsection{Setup for Llama-3-8b}
\label{sec-append-training-details-llama}

We use Lora~\citep{hu2021lora} to fine-tune Llama-3-8b. The setting for Lora are list below:
\begin{itemize}
    \item lora\_alpha: 16
    \item lora\_dropout: 0.1
    \item r: 64
    \item bias: none
    \item task\_type: CAUSAL\_LM
\end{itemize}

For each iteration, the settings below are the same.
\begin{itemize}
    \item er\_device\_train\_batch\_size: 4
    \item gradient\_accumulation\_steps: 1
    \item optim: paged\_adamw\_32bit
    \item learning\_rate: 2e-4
    \item weight\_decay: 0.001
    \item fp16: False
    \item bf16: False
    \item max\_grad\_norm: 0.3
    \item max\_steps: -1
    \item warmup\_ratio: 0.03
    \item group\_by\_length: True
    \item lr\_scheduler\_type: constant
    \item report\_to: tensorboard
\end{itemize}

For num\_train\_epochs, the details are shown in \Cref{tb-llama-nepochs}.

\section{Case study}
\label{sec-append-case-study}

\begin{figure}[t!]
    \centering
    \includegraphics[width=\textwidth]{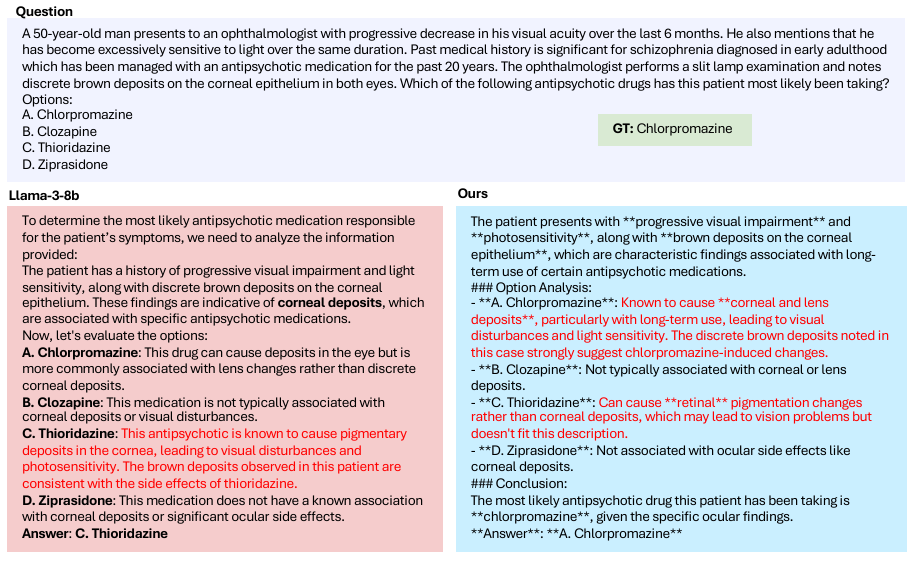}
    \caption{Case study on Llama-3-8b (1).}
    \label{fig-case-llama1}
\end{figure}

\begin{figure}[t!]
    \centering
    \includegraphics[width=\textwidth]{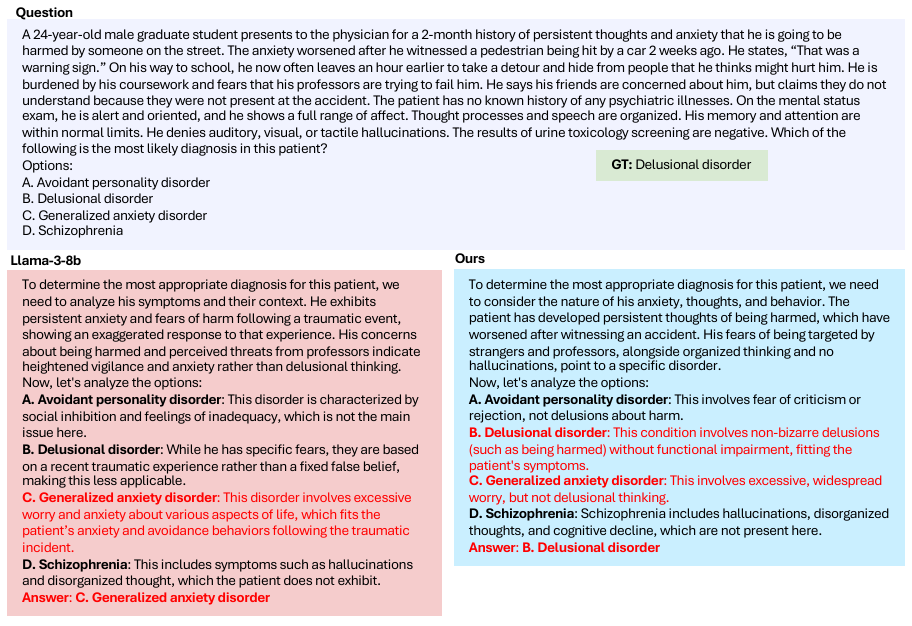}
    \caption{Case study on Llama-3-8b (2).}
    \label{fig-case-llama2}
\end{figure}

\begin{figure}[t!]
    \centering
    \includegraphics[width=\textwidth]{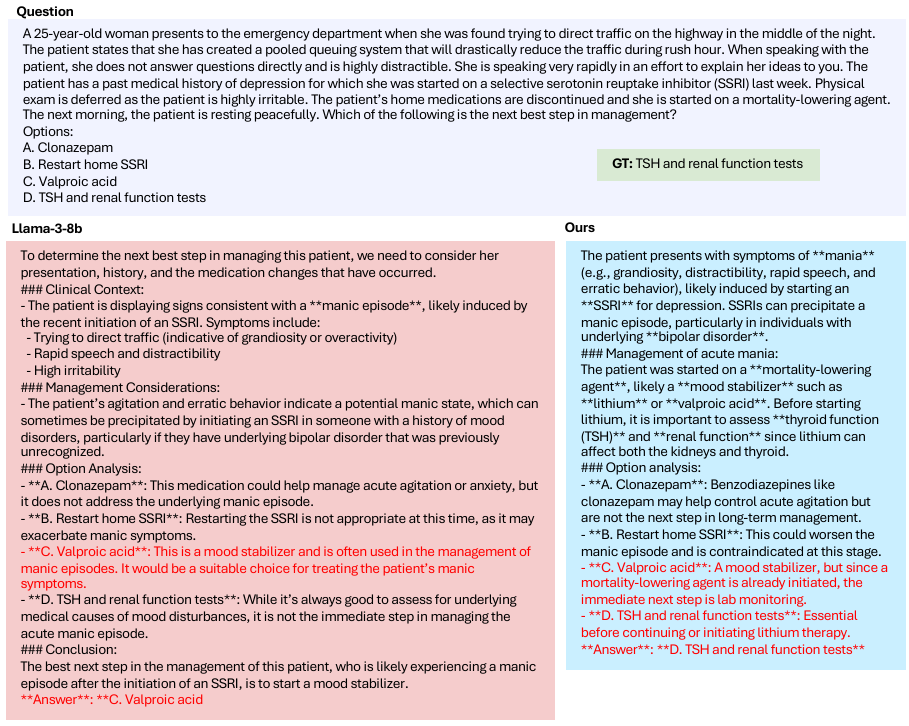}
    \caption{Case study on Llama-3-8b (3).}
    \label{fig-case-llama3}
\end{figure}

\begin{figure}[t!]
    \centering
    \includegraphics[width=\textwidth]{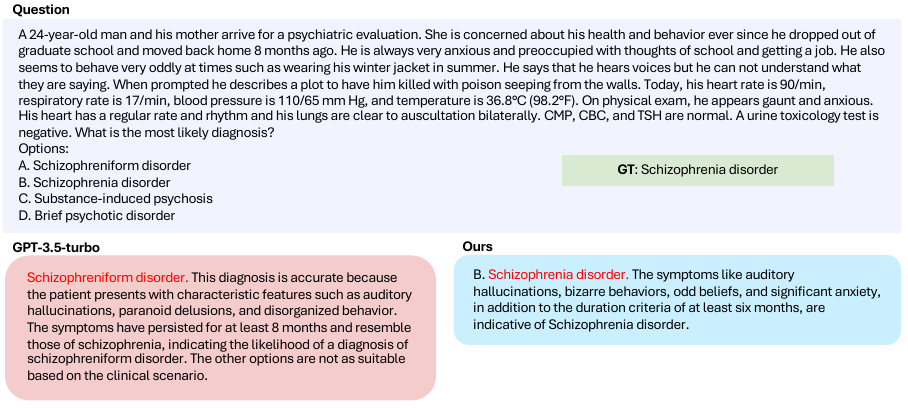}
    \caption{Case study on GPT-3.5-turbo (1).}
    \label{fig-case-gpt1}
\end{figure}

\begin{figure}[t!]
    \centering
    \includegraphics[width=\textwidth]{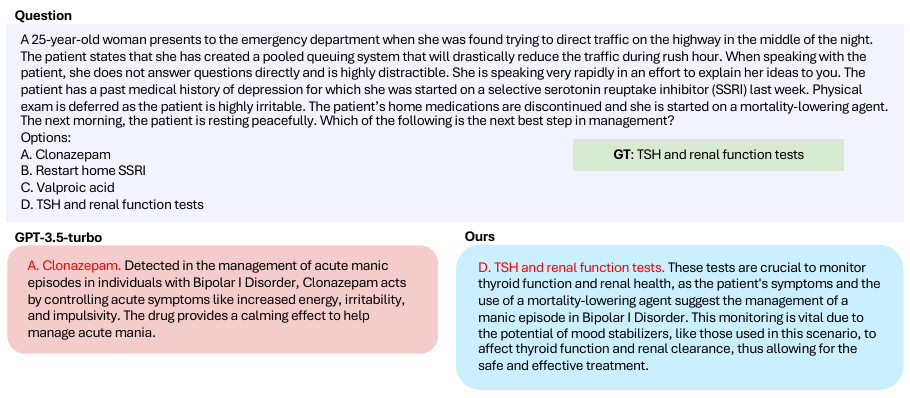}
    \caption{Case study on GPT-3.5-turbo (2).}
    \label{fig-case-gpt2}
\end{figure}

\begin{figure*}[t!]
    \centering
    \includegraphics[width=\textwidth]{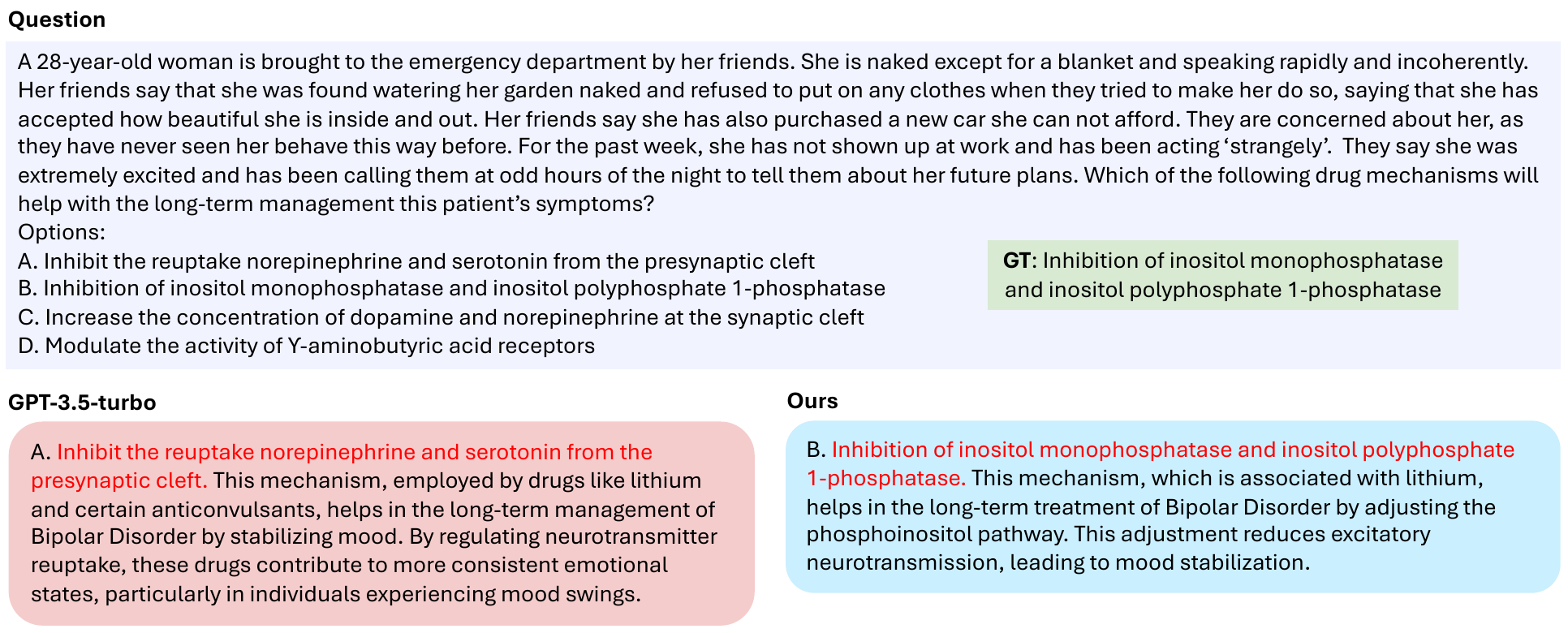}
    \caption{Case study on GPT-3.5-turbo (3).}
    \label{fig-case-gpt}
\end{figure*}

\Cref{fig-case-llama1,fig-case-llama2,fig-case-llama3,fig-case-gpt1,fig-case-gpt2} illustrate several cases comparing our models and the corresponding baseline models. Our models accurately answer the medical questions, while the base models provide incorrect responses. This discrepancy arises because the data generated during the patient-therapist interactions contains valuable medical knowledge, which aids in the analysis and formulation of the answer.

\end{document}